\documentclass{article}

\PassOptionsToPackage{numbers, compress}{natbib}
\usepackage[preprint]{neurips_2020} %
\usepackage[utf8]{inputenc} %
\usepackage[T1]{fontenc}    %
\usepackage{hyperref}       %
\usepackage{url}            %
\usepackage{booktabs}       %
\usepackage{amsfonts}       %
\usepackage{nicefrac}       %
\usepackage{microtype}      %
\usepackage{xcolor}

\usepackage{subfig}
\usepackage{pifont}
\usepackage{graphicx}
\graphicspath{{figure/}}
\usepackage{enumitem}
\usepackage{wrapfig}
\usepackage{lipsum}
\usepackage{mwe}

\newif\ifdraft
\draftfalse

\newcommand{\skdt}{sparse Kendall-Tau}
\newcommand{\pproxy}{$P_{proxy}$}
\newcommand{\fproxy}{$f_{proxy}$}
\newcommand{\pws}{$P_{ws}$}
\newcommand{\fws}{$f_{ws}$}

\newcommand{\mypara}[1]{\vspace{1mm}\noindent\textbf{#1}}

\title{
	How to Train Your Super-Net: An Analysis of Training Heuristics in Weight-Sharing NAS
}

\author{
	Kaicheng Yu\thanks{Correspondence to: kaicheng.yu@epfl.ch} 
	\thanks{EPFL-CVLab, \textsuperscript{$\ddagger$} Intel Labs, \textsuperscript{$\mathsection$}ClearSpace} 
	\And
	Rene Ranftl\textsuperscript{$\ddagger$}
	\And
	Mathieu Salzmann\textsuperscript{$\dagger$$\mathsection$}
}

\begin{document}
	
	\maketitle
	
	\begin{abstract}

Weight sharing promises to make neural architecture search (NAS) tractable even on commodity hardware.
Existing methods in this space rely on a diverse set of heuristics to design and train the shared-weight backbone network, a.k.a. the super-net. Since heuristics and hyperparameters substantially vary across different methods, a fair comparison between them can only be achieved by systematically analyzing the influence of these factors. In this paper, we therefore provide a systematic evaluation of the heuristics and hyperparameters that are frequently employed by weight-sharing NAS algorithms. Our analysis uncovers that some commonly-used heuristics for super-net training negatively impact the correlation between super-net and stand-alone performance, and evidences the strong influence of certain hyperparameters and architectural choices. Our code and experiments set a strong and reproducible baseline that future works can build on.
\end{abstract}

	\section{Introduction}
\label{sec:intro}

Neural architecture search~(NAS) has received growing attention in the past few years, yielding state-of-the-art performance on several machine learning tasks~\citep{liu2019autodeeplab,wu_fbnet:_2018,chen2019detnas,ryoo2020assemblenet}. One of the milestones that led to the popularity of NAS is weight sharing~\citep{Pham2018,Liu2018darts}, which, by allowing all possible network architectures to share the same parameters, has reduced the computational requirements from thousands of GPU hours to just a few. 

\begin{figure}
    \centering
    \resizebox{0.75\linewidth}{!}{
    \includegraphics{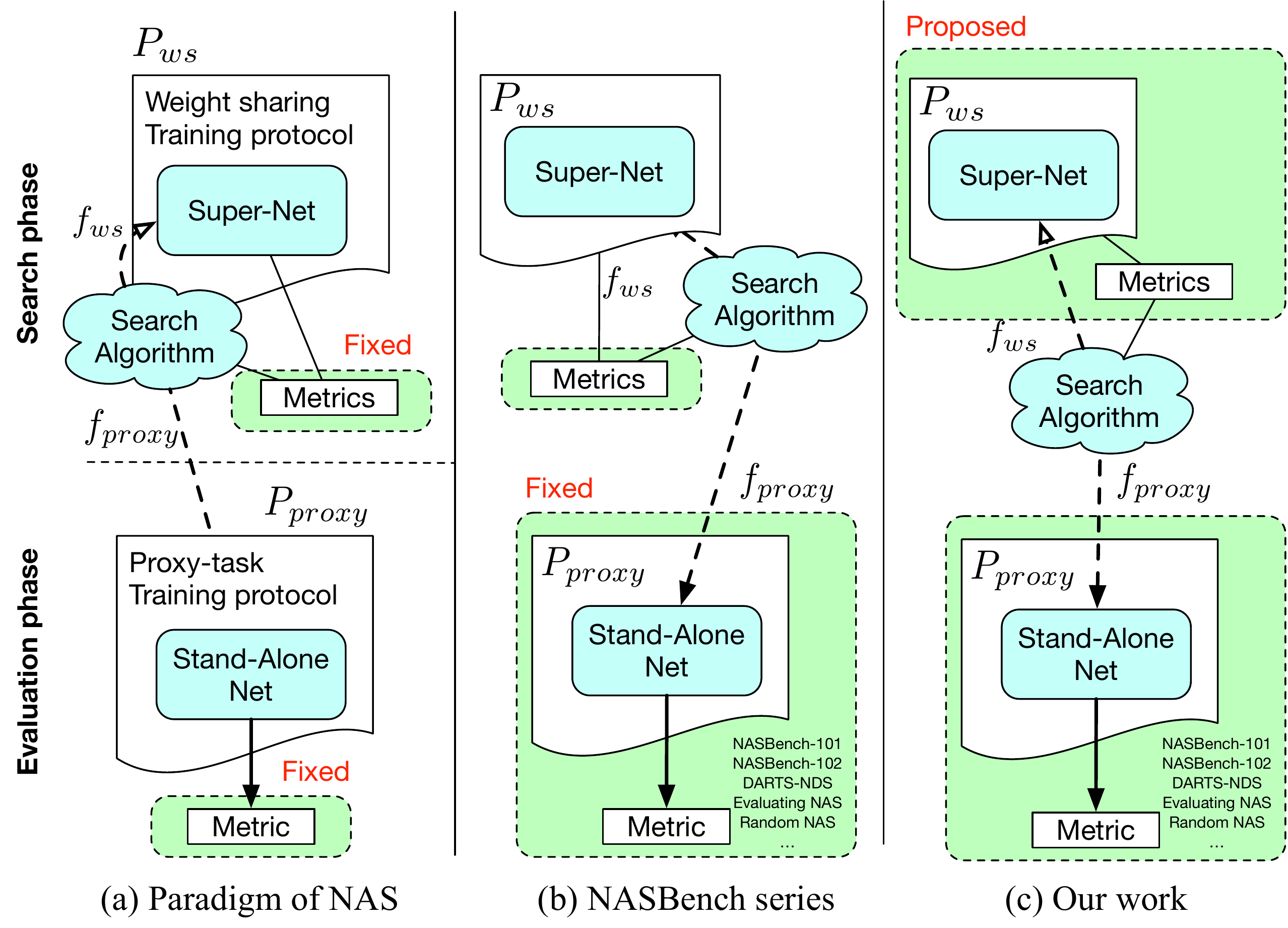}
    }
    \vspace{-.3cm}
    \caption{
    \textbf{WS-NAS benchmarking.}
    Green blocks indicate which aspects of NAS are benchmarked in different works.
    \textbf{(a)} 
    Early works fixed and compared the metrics on the proxy task, which doesn't allow for a holistic comparison between algorithms.
    \textbf{(b)} The NASBench benchmark series partially alleviates the problem by sharing the stand-alone training protocol and search space across algorithms. However, the design of the weight-sharing search space and training protocol is still not controlled. 
    \textbf{(c)} We fill this gap by benchmarking existing techniques to construct and train the shared-weight backbone. We provide a controlled evaluation across three benchmark spaces. $P$ indicates a training protocol, and $f$ a mapping function from the search space to a neural network.
    }
    \vspace{-0.6cm}
    \label{fig:teaser}
\end{figure}

Figure~\ref{fig:teaser} shows the two phases that are common to weight-sharing NAS (WS-NAS) algorithms: the search phase, including the design of the search space and the search algorithm; and the evaluation phase, which encompasses the final training protocol on the proxy task.

While most works focus on developing a good sampling algorithm~\citep{cai2018proxyless,xie2018snas} or improving  existing ones~\citep{Zela2020Understanding,nayman2019xnas,li2019improving}, 
the resulting methods differ in many other factors when it comes to designing and training the shared-weight backbone network, i.e., the super-net. For example, the literature reports diverse learning hyper-parameter settings, variations of how batch normalization and dropout are used, different capacities for the initial layers of the network, and variations in the depth of the super-net. These factors increase the difficulty to perform a fair comparison of NAS algorithms, and thus hinder our understanding of the reasons for success and failure of different strategies in different contexts. 

In this paper, we 
advance this understanding by performing a systematic evaluation of the effectiveness of commonly-used super-net design and training heuristics. To this end, we leverage three benchmark search spaces, NASBench-101~\citep{ying2019bench},  NASBench-201~\citep{dong2020bench102}, and DARTS-NDS~\citep{radosavovic_network_2019}, for which the ground-truth stand-alone performance of a large number of architectures is available. 
We report the results of our extensive experiments according to two sets of metrics: i) metrics that directly measure the quality of the super-net, such as the widely-adopted super-net accuracy and a modified Kendall-Tau correlation between the searched architectures and their ground-truth performance, which we refer to as \skdt{}; ii)
proxy metrics
such as the ability to surpass random search and the stand-alone accuracy of the model found by the WS-NAS algorithm.

Our analysis reveals (i) the factors that have a strong influence on the final performance, and thus strongly reduce the discrepancies between different 
search
algorithms; (ii) that some commonly-used training heuristics negatively affect performance; (iii) that some factors believed to have a strong impact on performance in fact only have a marginal effect. Furthermore, our evaluation shows that some search spaces are more amenable to weight sharing than others, and that the commonly-used super-net accuracy metric 
has a low correlation with the final stand-alone performance of a searched model. 
We show that our \skdt{} metric has a significantly higher correlation to stand-alone performance, and is thus a better metric to evaluate the training of the super-net. 
Ultimately, our analysis allows us to improve the super-net design on NASBench-101, significantly increasing the \skdt{} from 0.22 to 0.46.

Altogether, our work is the first to systematically analyze the impact of the diverse factors of super-net design and training. We uncover the factors that are crucial to design a super-net, as well as the non-important ones. 
Our analysis allows us to construct a new baseline that consistently achieves state-of-the-art search results with a weight-sharing random search algorithm. 
We will release our code and trained models so as to provide a unified WS-NAS framework.

\section{Preliminaries and Related Work}
We first introduce the necessary concepts that will be used throughout the paper. As shown in Fig.~\ref{fig:teaser}(a), weight-sharing NAS algorithms consist of three key components: a search algorithm that samples an architecture from the search space in the form of an encoding, a mapping function \fproxy{} that maps the encoding into its corresponding neural network, and a training protocol for a proxy task \pproxy{} for which the network is optimized.

To train the search algorithm, one needs to additionally define the mapping function \fws{} that generates the shared-weight network. Note that the mapping \fproxy{} frequently differs from \fws{}, since in practice the final model contains many more layers and parameters so as to yield competitive 
results on the proxy task. After fixing \fws{}, a training protocol \pws{} is required to learn the super-net. 
In practice, \pws{} often hides factors that are %
critical for the final performance of an approach, such as hyper-parameter settings or the use of data augmentation strategies to achieve state-of-the-art performance~\citep{Liu2018darts,chu_fairnas:_2019,Zela2020Understanding}. Again, \pws{} may differ from \pproxy{}, which is used to train the architecture that has been found by the search. For example, our experiments reveal that the learning rate and the total number of epochs frequently differ due to the different training behavior of the super-net and stand-alone architectures.

Many strategies have been proposed to implement the search algorithm, such as
reinforcement learning~\citep{Zoph2017,Zoph2018}, evolutionary algorithms~\citep{Real2017,Miikkulainen2019,so2019evolved,Liu2018b,Lu2018}, gradient-based optimization~\citep{Liu2018darts,Xu2020PC-DARTS:,li2019improving}, Bayesian optimization~\citep{kandasamy2018neural,jin2019auto,zhou2019bayesnas,wang2020neural}, and separate performance predictors~\citep{Liu2018b,Luo2018}.
Until very recently, the common trend to evaluate NAS consisted of reporting the searched architecture's performance on the proxy task~\citep{xie2018snas,real2018regularized,ryoo2020assemblenet}.
This, however, hardly provides real insights about the NAS algorithms themselves, because of the many 
components involved in them.
Many factors that differ from one algorithm to another can 
influence the performance. In practice, the literature even commonly compares NAS methods that employ different protocols to train the final model. 

\citet{li2019random} and \citet{yu2020evalnas} were the first to systematically compare different algorithms with the same %
settings for the proxy task and using several random initializations. Their surprising results revealed that many NAS algorithms produce architectures that do not significantly outperform a randomly-sampled architecture. 
\citet{Yang2020NAS} highlighted the importance of the training protocol \pproxy{}. They showed that optimizing the training protocol can improve the final architecture performance on the proxy task by three percent on CIFAR-10~\citep{Krizhevsky09}. This non-trivial improvement can be achieved regardless of the chosen sampler, which provides clear evidence for the importance of unifying the protocol to build a solid foundation for comparing NAS algorithms. 
In parallel to this line of research, the recent series of ``NASBench" works~\citep{ying2019bench,Zela2020NAS-Bench-1Shot1,dong2020bench102} proposed to benchmark NAS approaches by providing a complete, tabular characterization of the performance of every architecture in a given search space. This was achieved by training every realizable stand-alone architecture using a fixed protocol \pproxy{}. Similarly, other works proposed to provide a partial characterization by sampling and training a sufficient number of architectures in a given search space using a fixed protocol~\cite{radosavovic_network_2019,Zela2020Understanding,wang2020neural}.

While recent advances for systematic evaluation are promising, no work has yet thoroughly studied the influence of the super-net training protocol \pws{} and of the mapping function \fws{}. 
Previous works~\citep{Zela2020Understanding,li2019random} have performed hyper-parameter tuning to evaluate their own algorithms, and focused only on a few hyper-parameters.
This is the gap we fill here by benchmarking different choices of \pws{} and \fws{} that can be found in the WS-NAS literature. As will be shown in our experiments, this allows us to perform a fair comparison
of existing techniques across three benchmark datasets and to provide guidelines on how to train your super-net.

\section{Evaluation Methodology}

\label{sec:method-factors}
\begin{wrapfigure}{r}{0.5 \linewidth}
    \centering
    \vspace{-0.3cm}
    \resizebox{\linewidth}{!}{
    \includegraphics{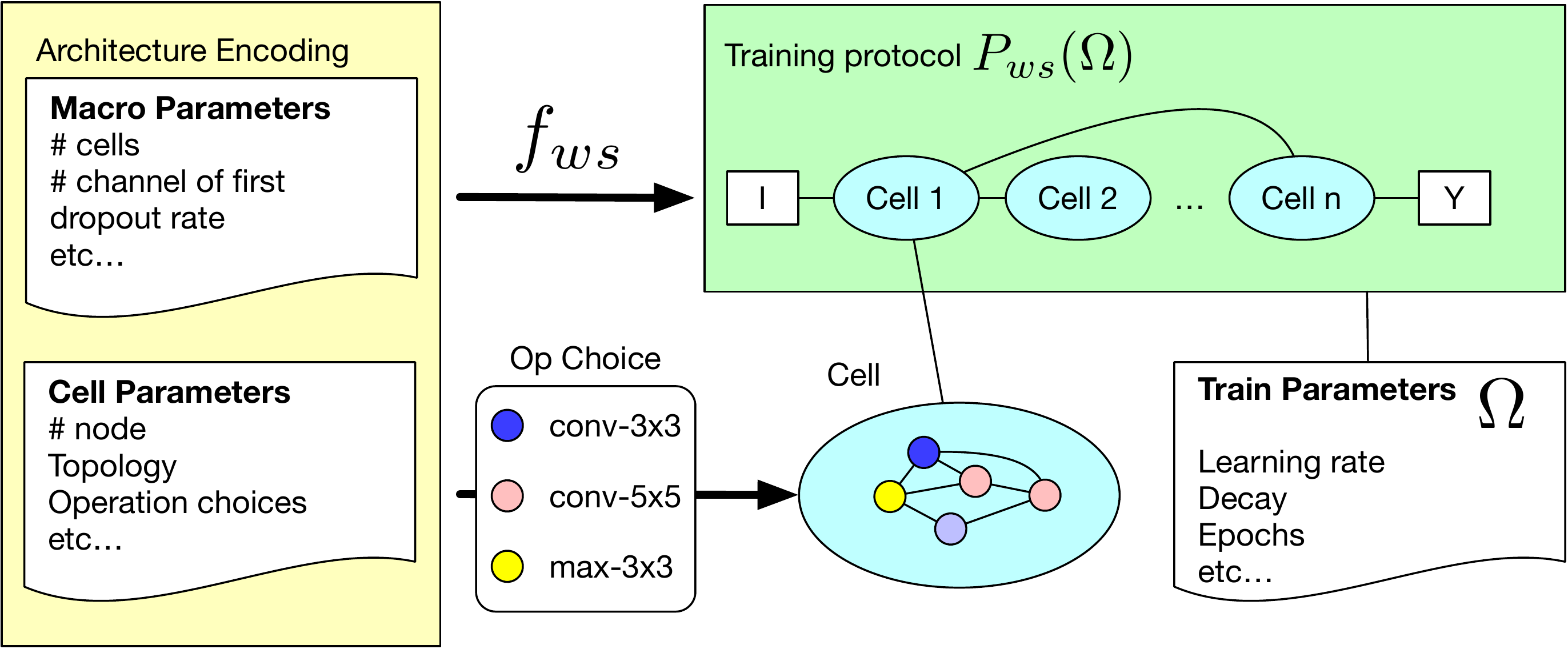}
    }\vspace{-0.1cm}
    \caption{\textbf{Constructing a super-net.} 
    }
    \vspace{-0.2cm}
    \label{fig:construction} 
\end{wrapfigure}
Our goal is to evaluate the influence of the super-net mapping $f_{ws}$ and weight-sharing training protocol $P_{ws}$.
As shown in Figure~\ref{fig:construction}, $f_{ws}$ translates an architecture encoding, which typically consists of a discrete number of choices or parameters, into a neural network. Based on a well-defined mapping, the super-net is a network in which every sub-path has a one-to-one mapping with an architecture encoding~\citep{Pham2018}. Recent works~\citep{Xu2020PC-DARTS:,li2019improving,ying2019bench} separate the encoding into \emph{cell parameters}, which define the basic building blocks of a network, and \textit{macro parameters}, which define the way cells are assembled into a complete architecture.

\mypara{Weight-sharing mapping $f_{ws}$.}
To make the search space manageable, all cell and macro parameters are fixed during the search, except for the topology of the cell and its possible operations. However, the exact choices for each of these fixed factors differ between algorithms and search spaces. We 
report the
common factors in the left 
part of Table~\ref{tab:factors}. They include various implementation choices, e.g., the use of convolutions with a dynamic number of channels~(Dynamic Conv),  super-convolutional layers that support dynamic kernel sizes (OFA Kernel)~\citep{Cai2020Once}, weight-sharing batch-normalization~(WSBN) that tracks independent running statistics and affine parameters for different incoming edges~\citep{Luo2018}, and path and global dropout~\citep{Pham2018,Luo2018,Liu2018darts}. They also include the use of low-fidelity estimates~\citep{elsken2019neural} to reduce the %
complexity of super-net training,
e.g., by reducing the number of layers~\citep{Liu2018darts} and channels~\citep{Yang2020NAS,chen2019pdarts}, the portion of the training set used for super-net training~\citep{Liu2018darts}, 
or the batch size\citep{Liu2018darts,Pham2018,Yang2020NAS}. 

\begin{table}[t]
    \centering
    \begin{minipage}[t]{0.48\linewidth}
    \centering
    \caption{\bf{Summary of factors}}
    \resizebox{\linewidth}{!}{
\begin{tabular}{cc|cc}
\toprule
 \multicolumn{2}{c}{WS Mapping $f_{ws}$} & \multicolumn{2}{c}{WS Protocol $P_{ws}$}  \\
\cmidrule{1-2} \cmidrule{3-4}
 \textit{implementation} & \textit{low fidelity} &\textit{hyperparam.} & \textit{sampling} \\
\midrule
 Dynamic Conv  & $\#$ layer & batch-norm & FairNAS  \\
OFA Conv & train portion & learning rate & Random-NAS \\
WSBN &  batch size & epochs & Random-A \\
Dropout & $ \# $ channels  & weight decay &  \\
Op on Node/Edge &&&\\
\bottomrule
\phantom{1}\\
\phantom{1}\\
\end{tabular}
} 
\label{tab:factors}
    \end{minipage}
    \hfill
    \begin{minipage}[t]{0.48\linewidth}
        \centering
    \caption{\bf{Search Spaces}.}
    \resizebox{\linewidth}{!}{
\begin{tabular}{l |ccc}
\toprule
& NASBench-101 & NASBench-201 & DARTS-NDS \\
\midrule
$\#$ Arch. & 423,624  & 15,625 & 5,000 \\
$\#$ Op. & 3 & 5 & 8 \\
Channel & Dynamic & Fix & Fix \\
Optimal & Global & Global & Sample \\
Nodes=($n$) & 5 & 4 & 4 \\
Param. & $O(n)$ & $O(n)$ - $O(n^2)$ & $O(n)$ - $O(n^2)$ \\
Edges & $O(n^2)$ & $O(n^2)$ & $O(n)$ \\
Merge & Concat. & Sum & Sum \\
\bottomrule
\end{tabular}}
\label{tab:search-space}
    \end{minipage}
\vspace{-0.6cm}
\end{table}

\mypara{Weight-sharing protocol $P_{ws}$.}
Given a mapping $f_{ws}$, different training protocols $P_{ws}$ can be employed to train the super-net. Protocols can differ in the training hyper-parameters and the sampling strategies they rely on. We will evaluate the different hyper-parameter choices listed in the right part of Table~\ref{tab:factors}. This includes the initial learning rate, the hyper-parameters of batch normalization, the total number of training epochs, and the amount of weight decay.

We restrict the search algorithm to the uniformly random sampling approach of~\citet{cai2018proxyless}, which is also known as single path one shot (SPOS)~\citep{guo_single_2019} or Random-NAS~\citep{li2019random}. 
The reason for this choice is that Random-NAS is equivalent to the initial state of many search algorithms~\citep{Liu2018darts,Pham2018,Luo2018}, some of which even 
freeze the sampler training so as to use random sampling
to warm-up the super-net~\citep{Xu2020PC-DARTS:,dong2020bench102}.
We additionally include two variants of Random-NAS: 1) As pointed out by~\citet{ying2019bench}, two super-net architectures might be topologically equivalent in the stand-alone network by simply swapping operations. We thus include architecture-aware random sampling that ensures equal probability for unique architectures~\citep{yu2020evalnas}. We name this variant Random-A; 
2) We evaluate a variant called FairNAS \citep{chu_fairnas:_2019}, which ensures that each operation is selected with equal probability during super-net training. Although FairNAS was designed for a search space where only operations are searched but not the topology, we adapt it to our setting (see Appendix \ref{apdx:fairnas} for details). 

In our experiments, for the sake of reproducibility, we ensure that $P_{ws}$ and \pproxy{}, as well as $f_{ws}$ and \fproxy{}, are as close to each other as possible. For the hyper-parameters of $P_{ws}$, we cross-validate each factor following the order in Table~\ref{tab:factors}, and after each validation, use the value that yields the best performance in \pproxy{}. 
For all other factors, we change one factor at a time.

\mypara{Search spaces.}
We use three commonly-used search spaces, for which a large number of stand-alone architectures have been trained and evaluated on CIFAR-10~\citep{Krizhevsky09} to obtain their ground-truth performance. In particular, we use NASBench-101~\citep{ying2019bench}, which consists of $423,624$ architectures and is compatible with weight-sharing NAS~ \citep{yu2020evalnas,Zela2020NAS-Bench-1Shot1}; NASBench-201~\citep{dong2020bench102}, which contains more operations than NASBench-101 but fewer nodes; and DARTS-NDS~\citep{radosavovic_network_2019} for which a subset of 5000 models was sampled and trained in a stand-alone fashion. A summary of these search spaces and their properties is shown in Table~\ref{tab:search-space}. The search spaces differ in the number of architectures that have known stand-alone accuracy~(\# Arch.), the number of possible operations~(\# Op.), how the  channels are handled in the convolution operations~(Channel), where dynamic means that the number of super-net channels might change based on the sampled architecture, and the type of optimum that is known for the search space~(Optimal). We further provide the maximum number of nodes ($n$), excluding the input and output nodes, in each cell, as well as a bound on the number of shared weights (Param.) and edge connections (Edges). Finally, the search spaces differ in how the nodes aggregate their inputs if they have multiple incoming edges (Merge). 

\mypara{Metrics.} 
We define two groups of metrics to evaluate different aspects of a trained super-net (see Appendix~\ref{apdx:hyper-param} for more detail and evaluation of hyper-parameters). 
The first group of metrics directly evaluates the quality of the super-net.
The first metric in this group is the accuracy of the super-net on the proxy task. Concretely, we report the average accuracy of
200 architectures on the validation set of the dataset of interest. We will refer to this metric simply as \textit{accuracy}. It is frequently used \cite{guo_single_2019,chu_fairnas:_2019} to assess the quality of the trained super-net, but we will show later that it is in fact a poor predictor of the final stand-alone performance.

\begin{wrapfigure}{r}{0.5\linewidth}
    \centering
    \vspace{-0.4cm}
    \begin{minipage}{.7\linewidth}
    \resizebox{\linewidth}{!}{
    \includegraphics{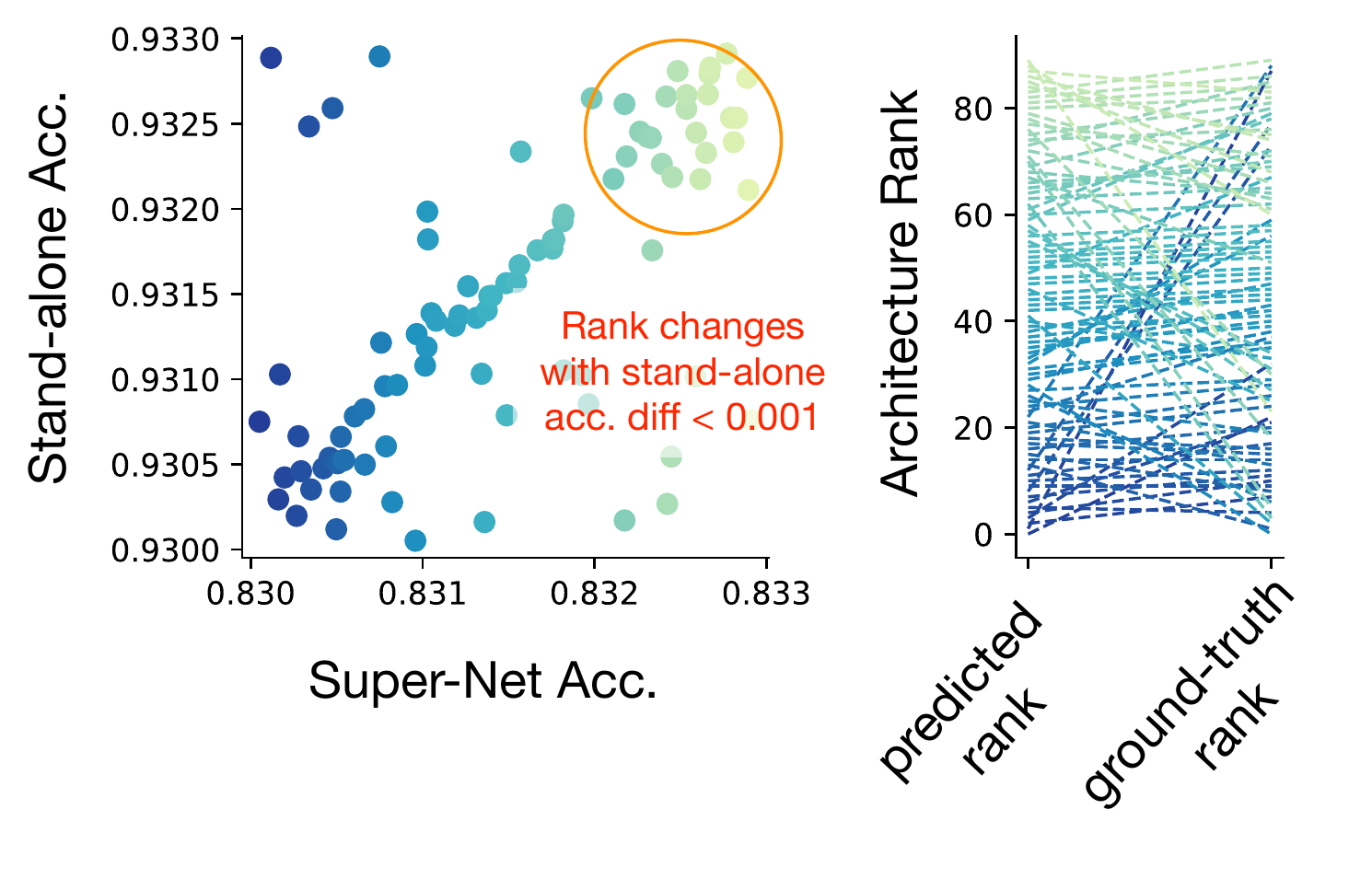}
    }
    \vspace{-0.5cm}
    \end{minipage}\hfill
    \begin{minipage}{0.3\linewidth}
    \vspace{-1.0cm}
    \resizebox{\linewidth}{!}{
    		\begin{tabular}{l|c}
			\toprule
			  &  Kendall\\
			  & Tau \\
			  \midrule
			  original  & 0.6444 \\
			  sparse &  0.8140  \\
			\bottomrule
    \end{tabular}
		}
    \end{minipage}
    \vspace{-0.35cm}
    \caption{\textbf{Kendall-Tau vs \skdt{}.} %
    Kendall-Tau is not robust when many architectures have similar performance (e.g. $\pm 0.1\%$). Minor performance differences can lead to large perturbations in the ranking. %
    Our \skdt{} alleviates this by dismissing minor differences in performance before constructing the ranking. 
    } 
    \label{fig:sparse-kdt}
    \vspace{-0.2cm}
\end{wrapfigure}

We further define a novel super-net metric, which we name \textit{\skdt{}}. It is inspired by the Kendall-Tau metric used by \citet{yu2020evalnas} to measure the discrepancy between the ordering of stand-alone architectures and the ordering that is implied by the trained super-net. An ideal super-net should yield the same ordering of architectures as the stand-alone one and thus would lead to a high Kendall-Tau. However, Kendall-Tau is not robust; the ranking of the architectures might be affected by negligible performance differences, translating to a low Kendall-Tau (c.f. Fig.~\ref{fig:sparse-kdt}). To robustify this metric, we share the rank between two architectures if their stand-alone accuracies differ by less than a threshold ($0.1\%$ here). Since the resulting ranks are sparse, we call this metric \textit{\skdt{}} (s-KdT). More details can be found in Appendix~\ref{apdx:skdt}.

The metrics in the second group evaluate the search performance of a trained super-net. 
The first metric is the probability to surpass random search. 
Given the ground-truth rank $r$ of the best architecture found after $n$ runs and the maximum rank $r_{max}$, equal to the total number of architectures, the probability that the best architecture found is better than a randomly searched one 
is computed as $p = 1 - (1 - (r / r_{max}))^n$.
Finally, where appropriate, we report the stand-alone accuracy of the model that was found by the complete WS-NAS algorithm. Concretely, we randomly sample 200 architectures, select the 3 best models based on the super-net accuracy and query the ground-truth performance. We then take the mean of these architectures as stand-alone accuracy. Note that the same architectures are used to compute the \skdt{}.

	\section{Evaluation Results}
In this section, we empirically explore the impact of the factors summarized in Table~\ref{tab:factors} on WS-NAS across three different search spaces. 
Unless otherwise specified, we mainly rely on the \skdt{} to discuss the impact of each factor, and use final search performance as a reference only. The reasons behind this choice are analyzed in Section~\ref{sec:discussion}. We report the training details in Appendix ~\ref{apdx:training} and the complete numerical results for all settings in Appendix~\ref{apdx:allfactor}.

\subsection{Weight-sharing Protocol $P_{ws}$ -- Hyper-parameters}

For each search space, we start our experiments based on the original hyper-parameters used in 
\begin{wrapfigure}{r}{0.5 \linewidth}
    \centering
    \vspace{-0.4cm}
    \resizebox{\linewidth}{!}{
    \includegraphics{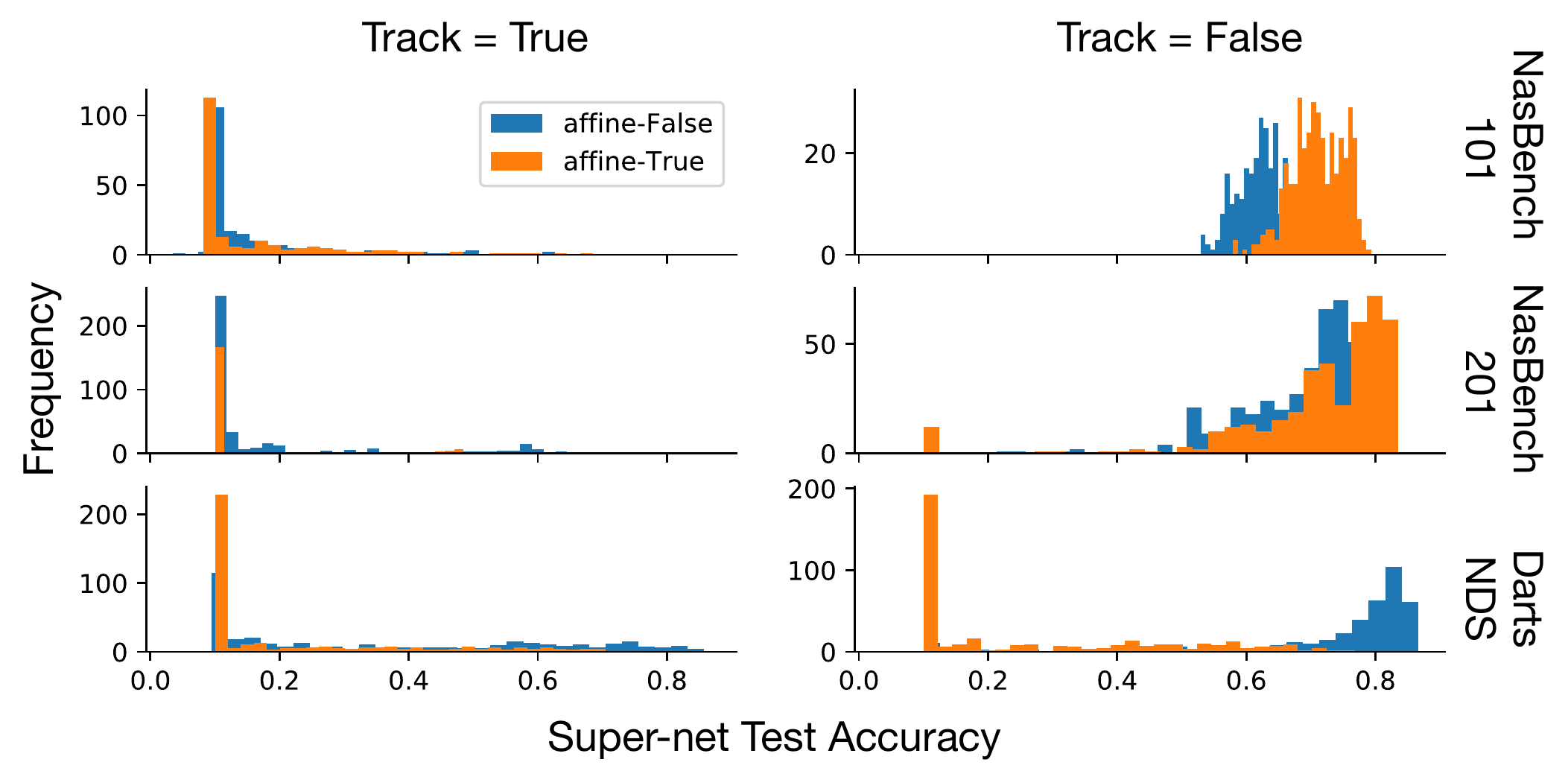}
    }
    \vspace{-0.6cm}
    \caption{\textbf{Validation of batch normalization.} 
    }
    \label{fig:bn}
    \resizebox{\linewidth}{!}{
    \includegraphics{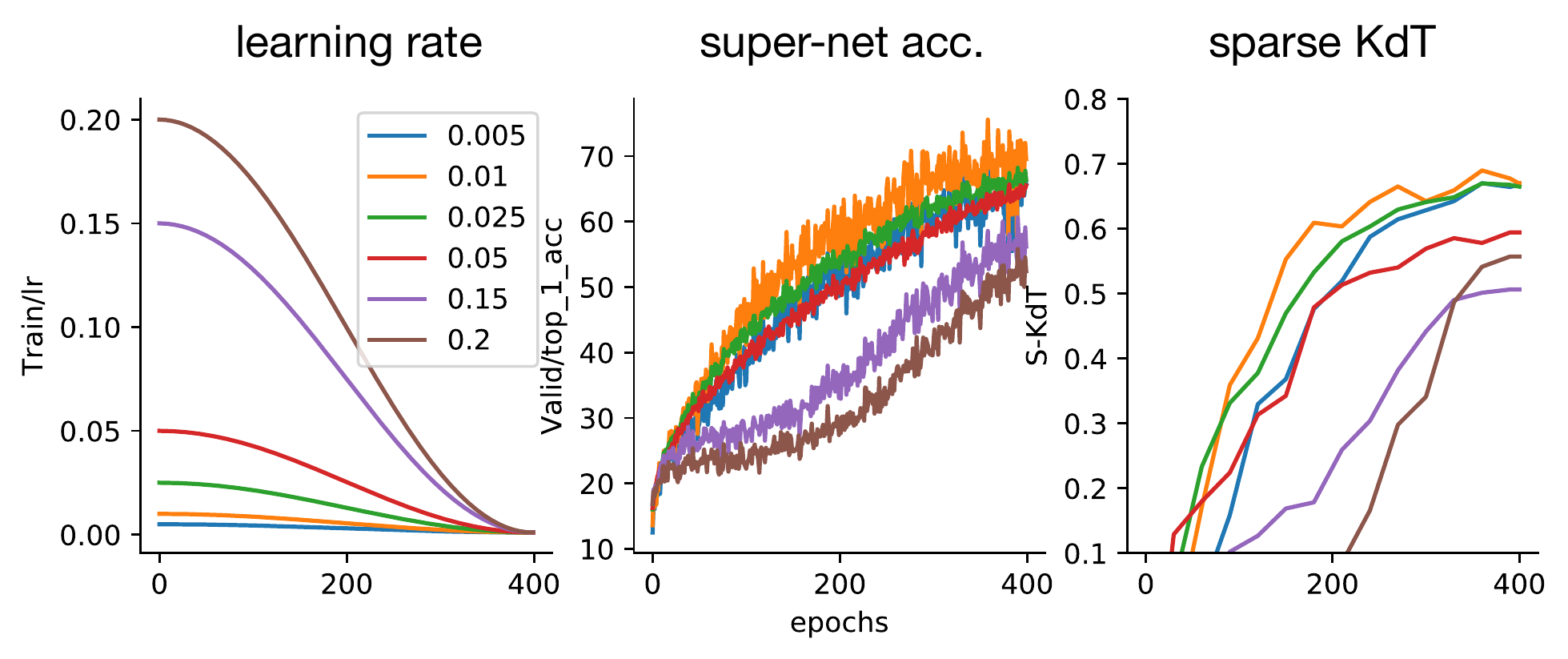}
    }
    \vspace{-0.6cm}
    \caption{\textbf{Learning rate} on NASBench-201 with 400 epochs. 
    }
    \vspace{-0.8cm}
    \label{fig:lr-nasbench201}
\end{wrapfigure}

stand-alone training. 
Because of the large number of hyper-parameters, we do not cross-validate all possible combinations. Doing so might further improve the performance.
We will use the parameters validated in this section in later experiments.

\mypara{Batch normalization.}
A fundamental assumption of batch normalization is that its input data follows a slowly changing distribution whose statistics can be tracked using a moving average during training. However, in WS-NAS, each node can receive wildly different inputs in every training iteration so that tracking the statistics is impossible.
As shown by Fig.~\ref{fig:bn}, 
using the tracked statistics severely hinders training and leads to many architectures having an accuracy of $\sim 10\%$, i.e., random predictions.  
This corroborates the discussion in~\citep{dong2020bench102}. We therefore do not track running statistics in the remaining experiments and only use mini-batch statistics. %
Our results also show that the choice of fixing vs. learning an affine transformation in batch normalization should match the stand-alone protocol \pproxy{}.
\mypara{Learning rate.} We observed that the learning rate has a critical impact on the training of the super-net. In the stand-alone protocol $P_{proxy}$, the learning rate is set to $0.2$ for NASBench-101, and to $0.1$ for NASBench-201 and DARTS-NDS. All protocols use a cosine learning rate decay.  Fig.~\ref{fig:lr-nasbench201} shows that super-net training requires lower learning rates than stand-alone training. This is reasonable as the loss in $P_{ws}$ can be thought of as the sum of millions of individual architecture losses. The same trend is shown for other datasets in Appendix~\ref{apdx:wsp}.  
We set the learning rate to 0.025 in the remaining experiments.

\mypara{Additional factors.} We report results of additional factors such as the number of training epochs and weight decay
in Appendix~\ref{apdx:wsp}. In summary, our experiments show that more training epochs 
positively influence super-net quality. The behavior of weight decay varies across datasets, and one cannot simply disable it as suggested by~\citet{nayman2019xnas}.

\subsection{Weight-sharing Protocol $P_{ws}$ -- Path Sampling}

\begin{figure}[t]
    \centering
    \vspace{-0.2cm}
    \resizebox{\linewidth}{!}{
    \includegraphics{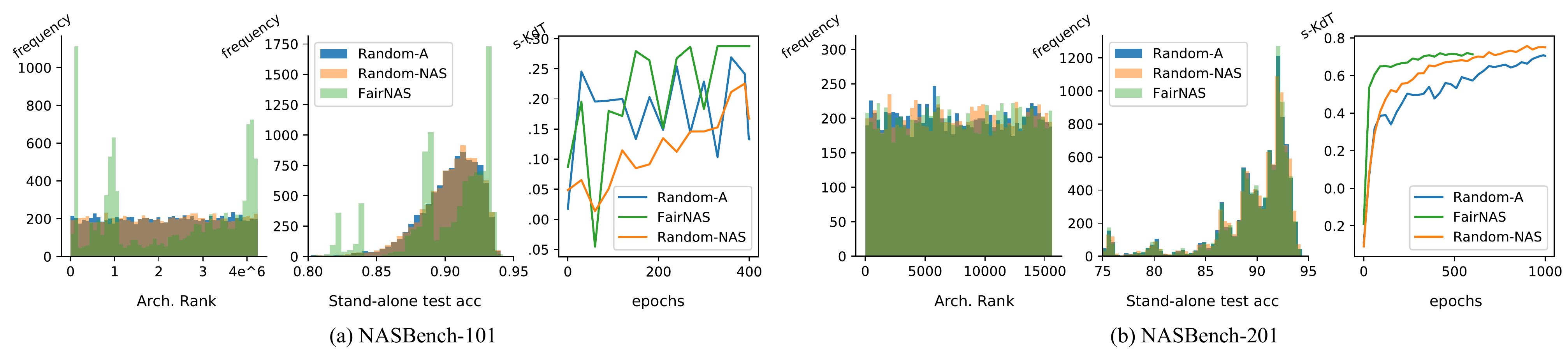}
    }
    \vspace{-0.5cm}
    \caption{
    \textbf{Path sampling comparison on NASBench-101 (a) and NASBench-201 (b).}
    We sampled 10,000 architectures using different samplers and plot histograms of the architecture rank and the stand-alone test accuracy. 
    We plot the s-KdT across the epochs.
    Results averaged across 3 runs.  
    }
    \vspace{-0.5cm}
    \label{fig:path-sample}
\end{figure}

With the hyper-parameters fixed, we now compare three path-sampling techniques. Since DARTS-NDS does not contain enough samples trained in a stand-alone manner, we only report results on NASBench-101 and 201. In Figure~\ref{fig:path-sample}, we show the sampling distributions of different approaches and the impact on the super-net in terms of \skdt{}. These experiments reveal that, on NASBench-101, uniformly randomly sampling one architecture, as in~\cite{li2019random,yu2020evalnas}, is strongly biased in terms of accuracy and ranking. This can be observed from the peaks around rank 0, 100,000, and 400,000. The reason is that a single architecture can have multiple  encodings, and uniform sampling thus  oversamples such architectures with
equivalent encodings.
FairNAS samples architectures more evenly and yields consistently better \skdt{}
values, albeit by a small margin.

On NASBench-201, the three sampling policies have a similar coverage. This is because, in NASBench-201, topologically-equivalent encodings were not pruned.
In this case, Random-NAS performs better than in NASBench-101, and FairNAS yields good early performance but quickly saturates. In short, using different sampling strategies might in general be beneficial, 
but we advocate for FairNAS in the presence of a limited training budget.

\subsection{Mapping $f_{ws}$ -- Lower Fidelity Estimates}

\begin{wrapfigure}{r}{0.33\linewidth}
    \centering
        \vspace{-.2cm}
    \resizebox{\linewidth}{!}{
    \includegraphics{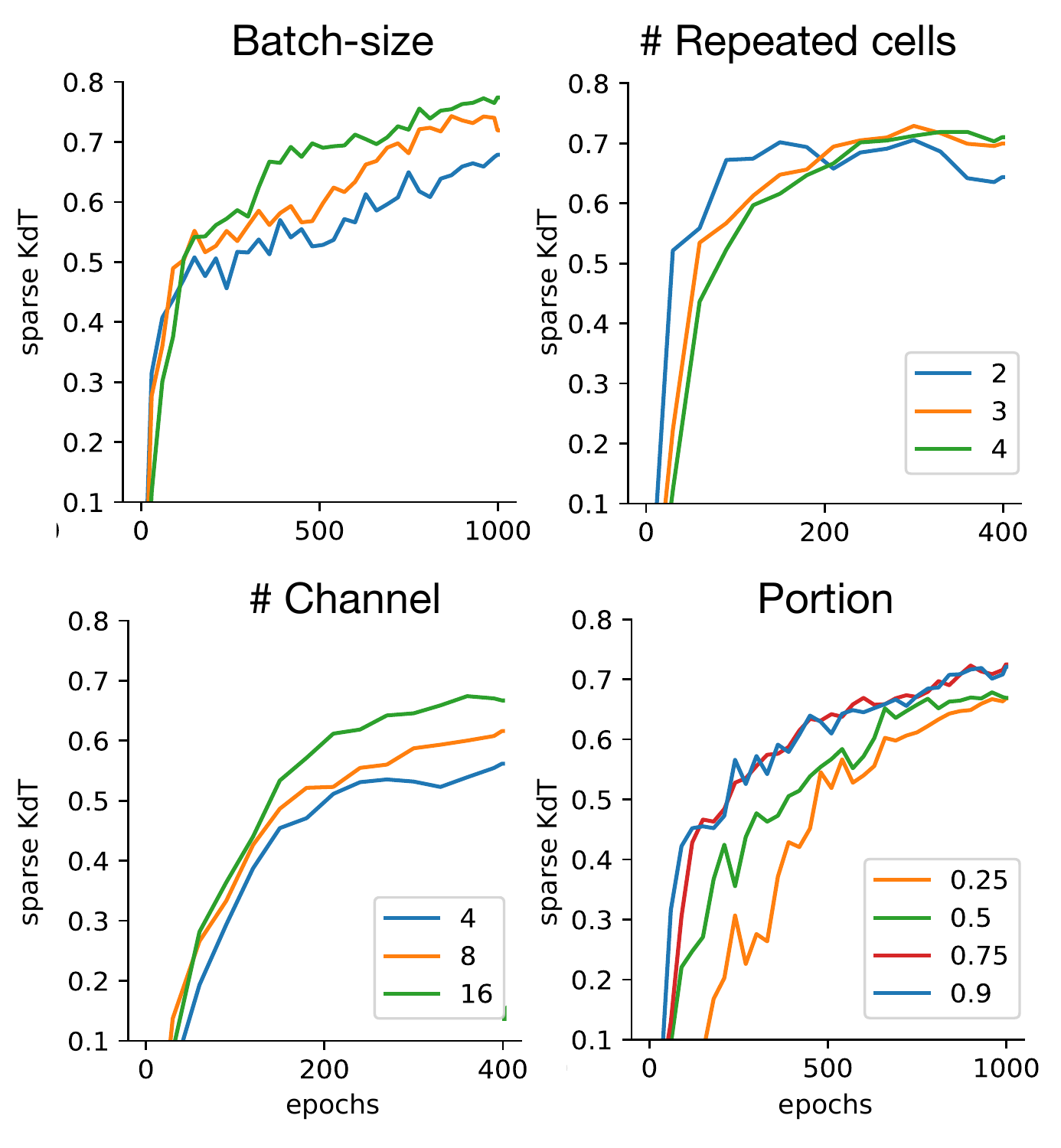}
    }
    \vspace{-0.7cm}
    \caption{\textbf{Low fidelity estimates} on NASBench-201. 
    }
    \vspace{-0.7cm}
    \label{fig:low-fidelity}
\end{wrapfigure}

Reducing memory foot-print and training time by proposing smaller super-nets has been an active research direction in WS-NAS, and the resulting super-nets are referred to as \textit{lower fidelity estimates}~\citep{elsken2019neural}. The impact of this approach on the super-net quality, however, has never been studied. We compare the influence of four commonly-used strategies in Figure~\ref{fig:low-fidelity}.

The most commonly-used approach to reduce memory requirement consists of decreasing the training batch size~\citep{Yang2020NAS}. Surprisingly, lowering the batch size from 256 to 64 has very limited impact on the super-net accuracy, but decreases the \skdt{} and the final searched model's performance. 

Another approach consists of decreasing the number of channels in the first layer~\citep{Liu2018darts}. 
This reduces the total number of parameters proportionally, since the number of channels in the consecutive layers
directly depends on the first one.
 
As can be seen in the corresponding plots, this decreases the \skdt{} from 0.7 to 0.5. 
By contrast, reducing the number of repeated cells~\citep{Pham2018,chu_fairnas:_2019} by 1 has little impact. Hence, to train a good super-net, one should avoid changes between $f_{ws}$ and \fproxy{}, but one can reduce the batch size by a factor > 0.5 and use only one repeated cell. 

The last lower-fidelity factor is the portion of training data that is  used~\citep{Liu2018darts,Xu2020PC-DARTS:}. Surprisingly, reducing the training portion only marginally decreases the \skdt{} for all three search spaces. On NASBench-201, keeping only 25\% of the CIFAR-10 dataset results in a 0.1 drop in \skdt{}. This explains why DARTS-based methods typically use only 50\% of the data to train the super-net but can still produce reasonable results.  

\subsection{Mapping $f_{ws}$ -- Implementation of the Layers}
We further validate the different implementations of the core layers in the mapping function $f_{ws}$. We analyze the dynamic channeling in detail and evaluate other mapping factors in Appendix~\ref{apdx:fws-other}. 

\begin{wraptable}{r}{0.5\linewidth}
    \centering
    \vspace{-0.4cm}
    \caption{ \textbf{Dynamic channels} on NASBench-101.} 
    \resizebox{0.95\linewidth}{!}{
        \begin{tabular}{ l|cccc}
        \toprule
        Type & Accuracy & S-KdT & P $>$ R & Final searched model \\
        \midrule
        Fixed       & 71.52 $\pm$ \phantom{0}6.94  & 0.22 & 0.546 & 91.79 $\pm$ 1.72 \\
        Shuffle     & 31.79 $\pm$ 10.90  & 0.17 & 0.391 & 90.58 $\pm$ 1.58 \\
        Interpolate & 57.53 $\pm$ 10.05 & 0.37 & 0.865 & 93.35 $\pm$ 3.27 \\
        \midrule
        Disable     & 76.95 $\pm$ \phantom{0}8.29 & 0.46 & 0.949 & 93.65 $\pm$ 0.73 \\
        \bottomrule
        \end{tabular}
    }
    \vspace{-0.2cm}
    \label{tab:dynamic-conv}
\end{wraptable}
\mypara{Dynamic channels.}
In NASBench-101, the output cell concatenates the feature maps from previous nodes. However, the concatenation has a fixed target size, which requires the number of output channels in the intermediate nodes to be dynamically adapted during super-net training. To model this, we initialize the super-net convolution weights so as to accommodate the largest possible number of channels $c_{max}$, and reduce it dynamically to $c$ output channels using one of the following heuristics: 1) Use a fixed chunk of weights, $[0:c]$~\citep{guo_single_2019}; 2) Shuffle the channels before applying 1)~\citep{zhang2018shufflenet}; 3) Linearly interpolate the $c_{max}$ channels into $c$ channels via a moving average across the neighboring channels.
The strategies are compared in Table~\ref{tab:dynamic-conv}. Shuffling the channels drastically degrades all metrics. Interpolation yields a lower super-net accuracy than using a fixed chunk, but improves the other metrics. Altogether, interpolation comes out as a more robust solution.

As dynamic channels are not used in NASBench-201, we evaluate the impact of removing this strategy on the results. To this end, we construct sub-spaces where the edge connections to the output nodes are fixed, which yields cells whose number of channels are all the same (see Appendix~\ref{apdx:nasbench-101} for more details). The results in Table~\ref{tab:dynamic-conv}  show a marked improvement, allowing us to reach the state-of-the-art in the NASBench-101 super-net design.

\section{Discussion and Conclusion}
\label{sec:discussion}

Let us now provide insight on how to evaluate a trained super-net and on the importance of different factors, and provide a concise set of rules to improve the training of a super-net. %

\mypara{Evaluation of the super-net.} 
The stand-alone performance of the architecture that is found by a NAS algorithm is clearly the most important metric to judge its merits. 
However, in practice, one cannot access this metric---we wouldn't need NAS if stand-alone performance was easy to query (the cost of computing stand-alone performance is discussed in Appendix~\ref{apdx:costFSA}). Furthermore stand-alone performance inevitably depends the sampling policy, and does not directly evaluate the quality of the super-net (see Appendix~\ref{apdx:fsa-vs-skt} for more details).
Consequently, it is important to 
rely on metrics that are well correlated with the final performance but can be queried efficiently. 
To this end, we collect all our experiments and plot the pairwise correlation between final performance, \skdt{}, and super-net accuracy. As shown in Figure~\ref{fig:supernet}, the super-net accuracy has a very low correlation with the final performance on NASBench-101 and DARTS-NDS. Only on NASBench-201 does it reach a  correlation of 0.52.
The \skdt{} yields a consistently higher correlation with the final performance than the super-net accuracy. This is the first concrete evidence that one should not focus too strongly on improving the super-net accuracy.
Note that the \skdt{} was computed using the same 200 architectures throughout the experiments. While the metric remains computationally heavy, it serves as a middle ground that is feasible to evaluate in real-world applications.

\begin{figure*}
    \centering
    \resizebox{\linewidth}{!}{
    \hspace{-1.cm}
    \includegraphics{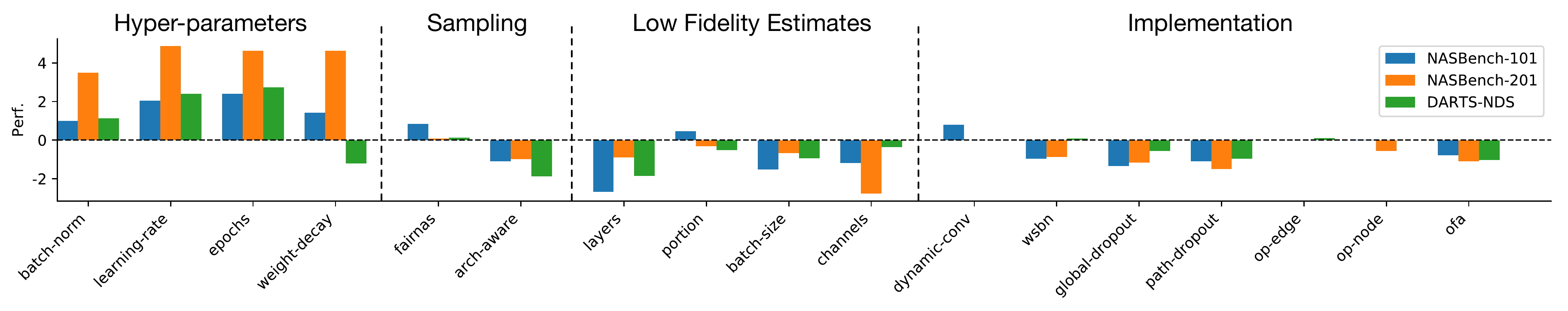}
    }
    \vspace{-0.7cm}
    \caption{\textbf{Influence of factors on the final model.} We plot the difference in percent between the searched model's performance with and without applying the corresponding factor. For the hyper-parameters of $P_{ws}$, the baseline is Random NAS, as reported in Table~\ref{tab:sota}. For the other factors, the baseline of each search space uses the best setting of the hyper-parameters. Each experiment was run at least 3 times. 
    }
       \vspace{-0.4cm}
    \label{fig:allfactors}
\end{figure*} 
\mypara{So, how should you train your super-net?}
Figure~\ref{fig:allfactors} summarizes the influence of the individual factors on the final performance.
It stands out that properly tuned hyper-parameters lead to the biggest improvements by far. Surprisingly, most other factors and techniques either have a hardly measurable effect or in some cases even lead to worse performance. The exceptions are FairNAS, which improves the stability of training and yields a small but consistent improvement, and to some degree the interpolation of dynamic convolutions.
Based on these findings, 
here is how you should train your super-net:
\begin{enumerate}[noitemsep,topsep=0pt,parsep=0pt,partopsep=0pt,leftmargin=12pt]
    \item Properly tune your hyper-parameters. Start from the hyper-parameters from the stand-alone protocol \pproxy{}, and rely on the order provided in Table~\ref{tab:factors}. 
\item Ensure a fair sampling. This refers to the super-net architecture space, not the stand-alone 
topologically-equivalent space, i.e., even if two architectures are equivalent when training them separately, we should treat them as two different architectures during super-net training.
\item Do not use super-net accuracy to judge the quality of your super-net. The \skdt{} has much higher correlation with the final search performance. 
\item Use low-fidelity estimates cautiously. Reducing the size of the training set moderately can nonetheless effectively reduce training time. 
\end{enumerate}

Finally, in Table~\ref{tab:sota}, we show that carefully controlling the relevant factors allows us to considerably improve the performance of Random-NAS. 
In short, thanks to our evaluation, we showed that simple Random-NAS together with an appropriate training protocol $P_{ws}$ and mapping function $f_{ws}$ yields results that are competitive to and sometimes even surpass state-of-the-art algorithms. We hope that our work will encourage the community to report detailed hyper-parameter settings to ensure that fair comparisons between NAS algorithms are possible. 
Our results provide a strong baseline upon which future work can build. %

\begin{figure}[!b]
    \centering
    \begin{minipage}[t]{0.48\linewidth}
        \centering
        \vspace{-0.4cm}
        \resizebox{\linewidth}{!}{
        \hspace{-1.cm}
        \includegraphics{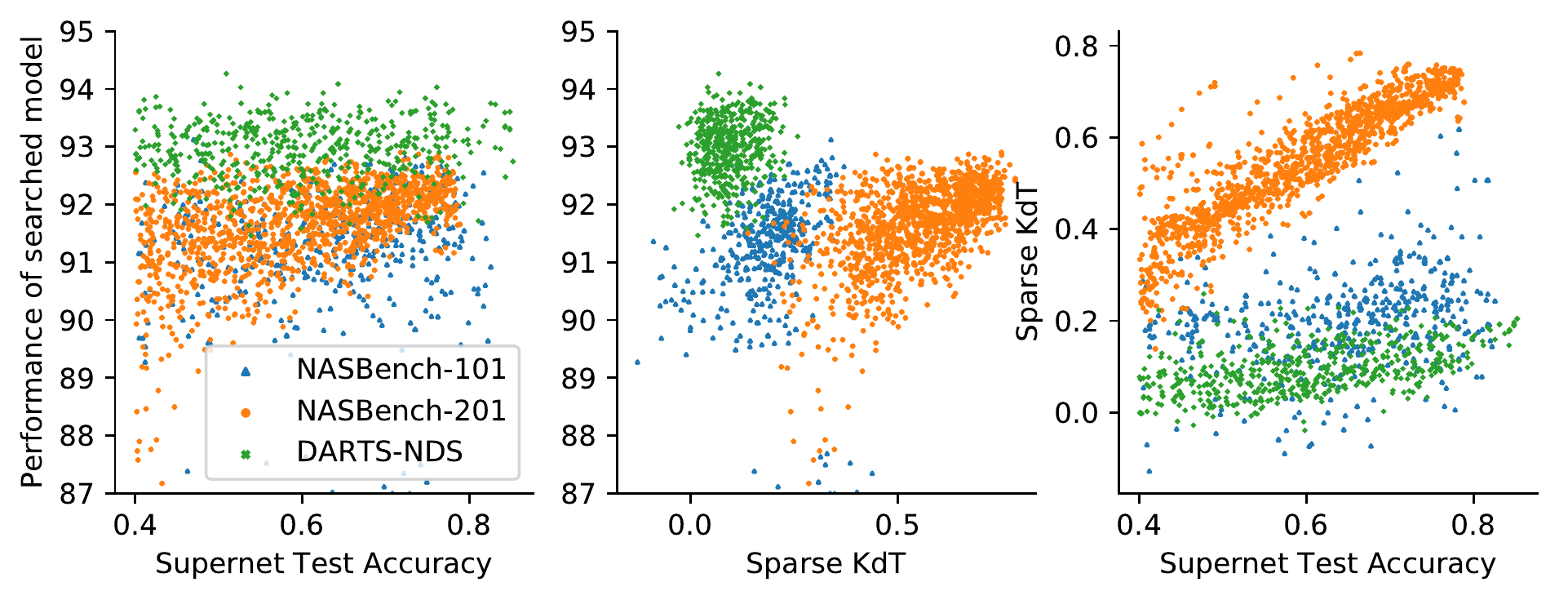}
        }
        \resizebox{0.8\linewidth}{!}{
        \begin{tabular}{l|ccc}
        Spearman Corr.     &  Acc. - Perf. & S-KdT - Perf. & Acc. - S-KdT \\
        \midrule
        NASBench-101 & 0.09  & 0.45 & 0.23 \\
        NASBench-201 & 0.52  & 0.55 & 0.94 \\
        DARTS-NDS    & 0.07  & 0.19 & 0.47 \\
        \end{tabular}}
        \captionof{figure}{\textbf{Super-net evaluation}. We collect all experiments across 3 benchmark spaces. \textbf{(Top)} Pairwise plots of super-net accuracy, final performance, and the \skdt{}. Each point corresponds to to one individual evaluation of the super-net. \textbf{(Bottom)} Spearman correlation coefficients between the metrics.
        }
        \label{fig:supernet}
    \end{minipage}
    \hfill
    \begin{minipage}[t]{0.48\linewidth}
        \centering
        \vspace{-0.4cm}
        \captionof{table}{
        \textbf{Search results on 3 search spaces.} 
        Results on NASBench-101 and NASBench-201 are taken from \citet{yu2020evalnas}, and \citet{dong2020bench102}. We report the mean ($\pm$ std) over 3 different runs. Note that NASBench-101 (n=7) in \citep{yu2020evalnas} is identical to our setting. Our new strategy significantly surpasses the random search baseline.
        }
        \resizebox{\linewidth}{!}{
        \begin{tabular}{l|cccc}
        \toprule
             Method &  NASBench & NASBench & DARTS & DARTS \\
             & 101 & 201 & NDS & NDS$^\star$ \\
             \midrule
             ENAS & 91.83 $\pm$ 0.42 & 54.30 $\pm$ 0.00  & 94.45& 97.11 \\
             DARTS-V2 & 92.21 $\pm$ 0.61 & 54.30 $\pm$ 0.00 & 94.79 & 97.37 \\
             NAO  & 92.59 $\pm$ 0.59 & - & - & 97.10 \\
             GDAS & - & 93.51 $\pm$ 0.13 & - & 96.23 \\
             \midrule 
              Random NAS & 89.89 $\pm$ 3.89 & 87.66 $\pm$ 1.69 & 91.33 & 96.74$^\dag$ \\
             Random NAS (Ours) & 93.12 $\pm$ 0.06 & 92.90 $\pm$ 0.14 & 94.26 $\pm$ 0.05 & 97.08 \\
             \bottomrule
             \multicolumn{5}{l}{$^{\dag}$Result took from \citet{li2019random}} \\
             \multicolumn{5}{l}{$^{\star}$Trained according to \citet{Liu2018darts} for 600 epochs.} \\
             \multicolumn{5}{l}{DARTS-V2~\citep{Liu2018darts}, ENAS~\citep{Pham2018},NAO~\citep{Luo2018}.} \\
             \multicolumn{5}{l}{Random-NAS~\citep{li2019random},GDAS~\citep{dong2019searching}} \\
        \end{tabular}
        }
        \label{tab:sota}
    \end{minipage}
\vspace{-0.6cm}
\end{figure}

	\section{Broader Impact}
	NAS has been drawing heated attention, particularly thanks to weight sharing, which makes it computationally tractable.
	One could thus think that we have reached the point where the industry can reliably exploit these results and discover the architectures best suited to its various needs.
	
	As a matter of fact, our interest in NAS was initiated by an industrial collaboration. However, we quickly found through our own experiments that NAS algorithms were hard to reproduce, and our industrial partner got discouraged. The work we present in this paper directly addresses this drawback. By providing the first systematic analysis of different technical choices and hyper-parameters in WS-NAS, our study explains why existing approaches often do not work well in practice: relevant aspects and parameters are simply not reported in publications. We thus expect that our effort will motivate researchers to more thoroughly explore all aspects of their work and to ensure that it is reproducible, which in turn will facilitate the deployment of NAS in the real world.
	
	We nonetheless acknowledge that the power consumption of NAS, even with weight sharing, remains high. This could lead to negative impact on the environment and contribute to global-warming. Yet, we hope that our guidelines will help the NAS community, academics and  practitioners,
	to reduce the number of unnecessary trials and thus save the planet by a baby step. 
	
	\section{Acknowlegement}
	Work done during an internship at Intel, and partially supported by the Swiss National Science Foundation.

\newpage
\appendix
\section{Methodology Details}

In this section, we provide some additional details about our methodology.
\subsection{Adaptation of FairNAS}
\label{apdx:fairnas}
\begin{wrapfigure}{r}{0.5\linewidth}
    \vspace{-0.5cm}
    \centering
    \resizebox{\linewidth}{!}{
    \includegraphics{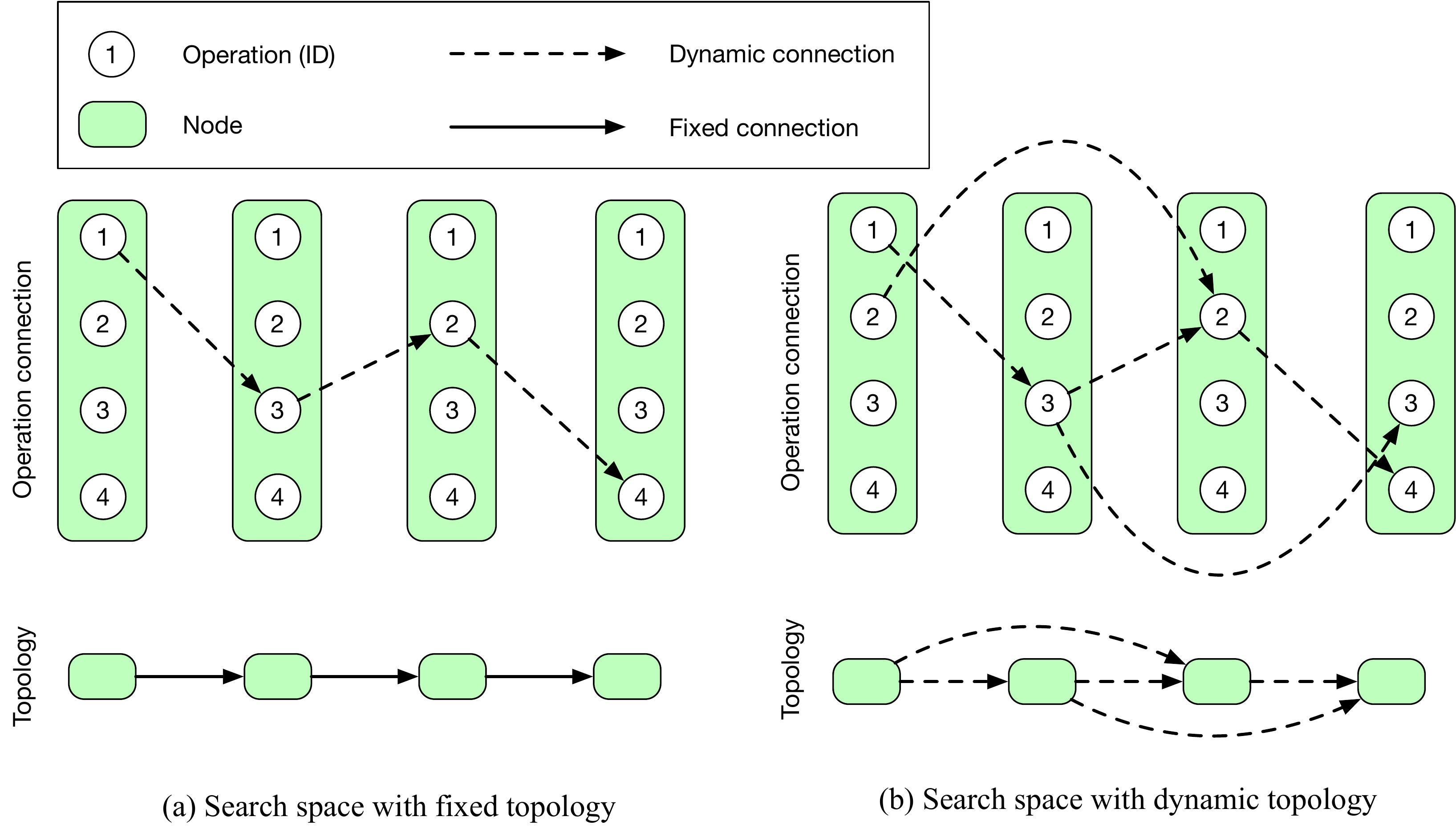}
    }
    \vspace{-0.5cm}
    \caption{Comparison between fixed and dynamic topology search spaces. }
    \label{fig:fairnas}
    \vspace{-0.3cm}
\end{wrapfigure}
Originally, FairNAS~\citep{chu_fairnas:_2019} was proposed in a search space with a fixed sequential topology, as depicted by Figure~\ref{fig:fairnas} (a), where every node is sequentially connected to the previous one, and only the operations on the edges are subject to change. However, our benchmark search spaces exploit a more complex dynamic topology, as illustrated in Figure~\ref{fig:fairnas} (b), where one node can connect to one or more previous nodes. 

Before generalizing to a dynamic topology search space, we simplify the original approach into a 2-node scenario: for each input batch, FairNAS will first randomly generate a sequence of all $o$ possible operations.
It then samples one operation at a time, computes gradients for the fixed input batch, and accumulates the gradients across the operations. Once all operations have been sampled, the super-net parameters are updated with the average gradients. 
This ensures that all possible paths are equally exploited . With this simplification, FairNAS can be applied regardless of the topology. For a sequential-topology search space, we repeat the 2-node policy for every consecutive 
node pair. Naturally, for a dynamic topology space, FairNAS can be adopted in a similar manner, i.e., one first samples a topology, then applies the 2-node strategy for all connected node pairs. Note that  
adapting
FairNAS increases the training time by a factor $o$. 

\section{Experimental Details}

Below, we provide additional details about our implementations and settings. We also report the best settings in Table~\ref{tab:best-setting}.

\begin{table*}[t]
    \centering
    \caption{Parameter settings that obtained the best searched results. }
    \resizebox{\textwidth}{!}{
    \begin{tabular}{l|ccccc|cccc|cccc|c}
    \toprule
        Search Space & \multicolumn{5}{c}{\textit{implementation}} & \multicolumn{4}{c}{\textit{low fidelity}} & \multicolumn{4}{c}{\textit{hyperparam.}} & \textit{sampling} \\
        \cmidrule{1-5} \cmidrule{6-9} \cmidrule{10-13} \cmidrule{14-15}
        & Dynamic Conv & OFA Conv & WSBN & Dropout & Op map & $\#$ layer & portion & batch-size & $\#$ channels & batch-norm & learning rate & epochs & weight decay &  \\
    \midrule
    NASBench-101 & Interpolation & N  & N & 0. & Node & 9 & 0.75 & 256 & 128 & Tr=F A=T & 0.025 & 400 & 1e-3 & FairNAS  \\
    NASBench-201 & Fix & N  & N & 0. & Edge & 5 & 0.9 & 128 & 16 &  Tr=F A=T & 0.025 & 1000 & 3e-3 & FairNAS  \\
    DARTS-NDS & Fix & N  & Y & 0. & Edge & 12 & 0.9 & 256 & 36 & Tr=F A=F & 0.025 & 400 & 0 & FairNAS  \\
    \bottomrule
    \multicolumn{15}{l}{For batch-norm, we report Track statistics (Tr) and Affine (A) setting with True (T) or False (F).}\\
    \multicolumn{15}{l}{For other notation, Y = Yes, N = No.}
    \end{tabular}
    }
    \label{tab:best-setting}
\end{table*}

\subsection{Sparse Kendall-Tau Hyper-parameters}
\label{apdx:hyper-param}
\mypara{Sparse Kendall-Tau threshold}.
This value should be chosen according to what is considered a significant improvement for a given task. For CIFAR-10, where accuracy is larger than 90\%, we consider a 0.1\% performance gap to be sufficient. For tasks with smaller state-of-the-art performance, larger values might be better suited.

\mypara{Number of architectures}.
In practice, we observed that the \skdt{} metric became stable and reliable when using at least $n=150$ architectures. We used $n=200$ in our experiments to guarantee stability and fairness of the comparison of the different factors.

\subsection{Sparse Kendall-Tau Implementation Details}
\label{apdx:skdt}
To compute the \skdt{} we need access to two quantities: 1) the performance of the sampled architectures based on the trained super-net; and 2) the associated ground-truth performances. For each architecture in 1), we compute the average top-1 accuracy over $n=3$ 
super-nets (that where trained with different random initialization) to improve the stability of the evaluation. We round the ground-truth top-1 accuracy to a precision of 0.1$\%$ for each sampled architecture to obtain the ground-truth performance 2). We then rank the architectures in 1) and 2) and compute the Kendall-Tau rank coefficient \cite{Kendall1938} between the two ranked lists. 

\subsection{Training Details}
\label{apdx:training}
We use PyTorch~\citep{pytorch} for our experiments. Since NASBench-101 was constructed in TensorFlow we implement a mapper that translates TensorFlow parameters into our PyTorch model. We exploit two large-scale experiment management tools, SLURM~\citep{slurm} and Kubernetes~\citep{k8s}, to deploy our experiments.
We use various GPUs throughout our project, including NVIDIA Tesla V100, RTX 2080 Ti, GTX 1080 Ti and Quadro 6000 with CUDA 10.1.
Depending on the number of training epochs, parameter sizes and batch-size, most of the super-net training finishes within 12 to 24 hours, with the exception of FairNAS, whose training time is longer, as discussed earlier. 
We split the data into training/validation using a 90/10 ratio for all experiments, except those involving validation on the training portion. Please consult our submitted code for more details.

\begin{figure}
    \centering
    \resizebox{0.75\linewidth}{!}{
    \includegraphics{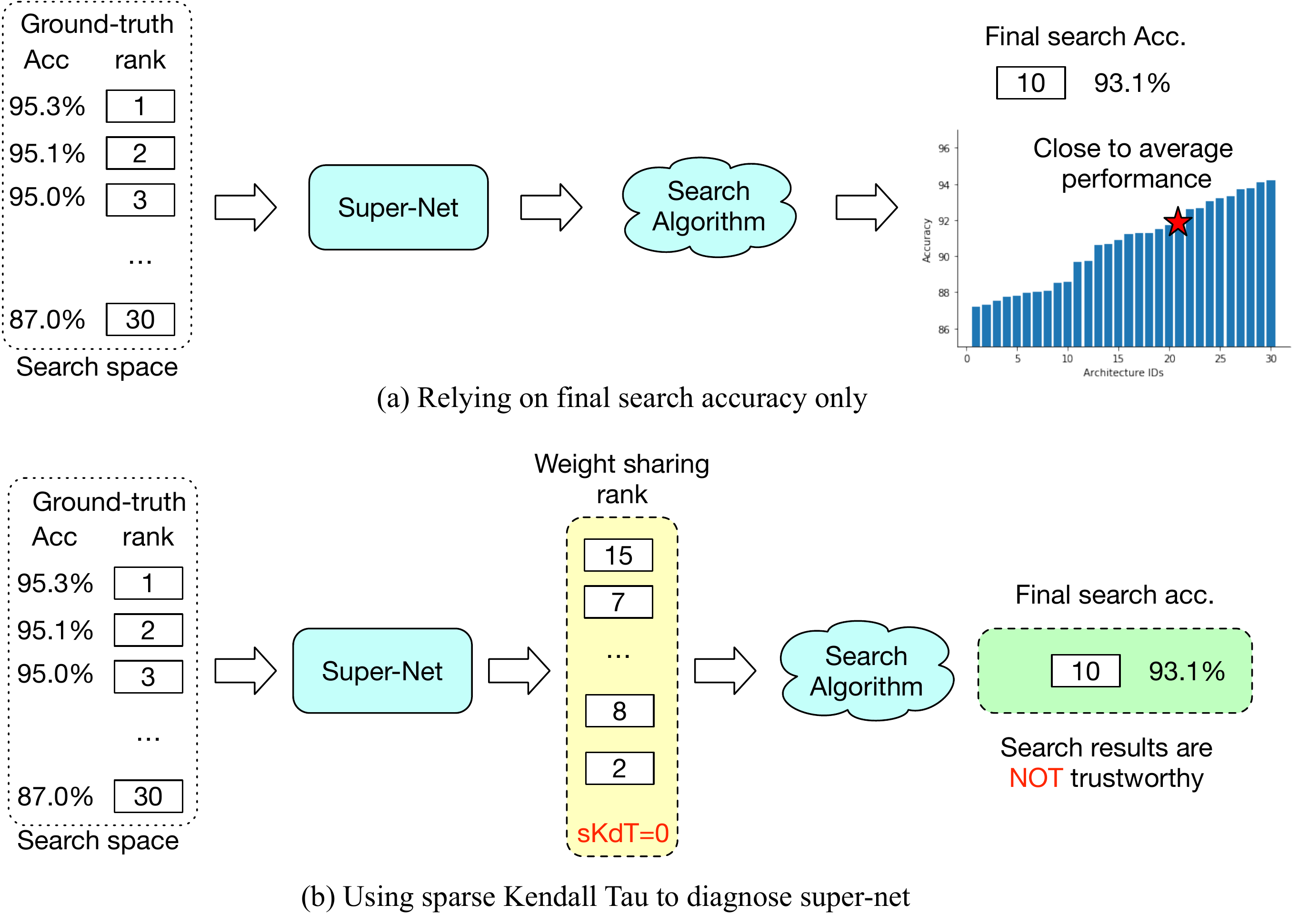}
    }
    \vspace{-0.3cm}
    \caption{\textbf{Comparing \skdt{} and final search accuracy.} Here, we provide a toy example to illustrate why one cannot rely on the final search accuracy to evaluate the quality of the super-net. Let us consider a search space with only 30 architectures, whose accuracy ranges from 95.3\% to 87\% on the CIFAR-10 dataset, and we run a search algorithm on top. \textbf{(a)} describes a common scenario: we run the search for multiple times, yielding a best architecture with 93.1\% accuracy. While this may seem good, it does not give any information about the quality of the search or the super-net. If we had full knowledge about the performance of every architecture in this space, we would see that this architecture is close to the average performance and hence no better than random. 
    In \textbf{(b)}, the \skdt{} allows us to diagnose this pathological case. A small \skdt{} implies that there is a problem with super-net training.
    }
    \label{fig:skdt-fsa}
\end{figure}
\subsection{Cost of Computing the Stand-alone Model Performance}
\label{apdx:costFSA}
Computing the final accuracy is more expensive than training the super-net. Despite the low-fidelity heuristics reducing the weight-sharing costs, training a stand-alone network to convergence has higher cost, e.g., DARTS searches for 50 epochs but trains from scratch for 600 epochs~\citep{Liu2018darts}. Furthermore, debugging and hyper-parameter tuning typically require training thousands of stand-alone models. Note that, as one typically evaluates a random subset of architectures to understand the design space~\citep{radosavovic_network_2019}, \skdt{} can be computed without additional costs. In any event, the budget for \skdt{} is bounded with $n$.

\subsection{Stand-alone Performance v.s. Sparse Kendall-Tau}
\label{apdx:fsa-vs-skt}
A common misconception is that the super-net quality is well reflected by stand-alone performance. Neither \skdt{} (sKT) nor final search accuracy (FSA) are perfect. Both are tools to measure different aspects of a super-net.
Below, we discuss this in more detail. 

Let us consider a completely new search space in which we have no prior knowledge about performance. As depicted by Figure~\ref{fig:skdt-fsa}, if we only rely on the FSA, the following situation might happen:
Due to the lack of knowledge, the ranking of the super-net is purely random, and the search space accuracy is uniformly distributed. When trying different settings, there will be 1 configuration that ‘outperforms’ the others in terms of FSA. However, this configuration will be selected by pure chance. By only measuring FSA, it is technically impossible to realize that the ranking is random. 
By contrast, if one measures the sKT (which is close to 0 in this example), an ill-conditioned super-net can easily be identified. In other words, purely relying on FSA could lead to pathological outcomes that can be avoided using sKT.

Additionally, FSA is related to both the super-net and the search algorithm. sKT allows us to judge super-net accuracy independently from the search algorithm. As an example, consider the use of a reinforcement learning algorithm, instead of random sampling, on top of the super-net. When observing a poor FSA, one cannot conclude if the problem is due to a poor super-net or to a poor performance of the RL algorithm. Prior to our work, people relied on the super-net accuracy to analyze the super-net quality. This is not a reliable metric, as shown in Fig.~9 in the main paper.  We believe that sKT is a better alternative. 

\begin{figure}[t]
    \centering
    \vspace{-0.2cm}
    \resizebox{\linewidth}{!}{
    \includegraphics{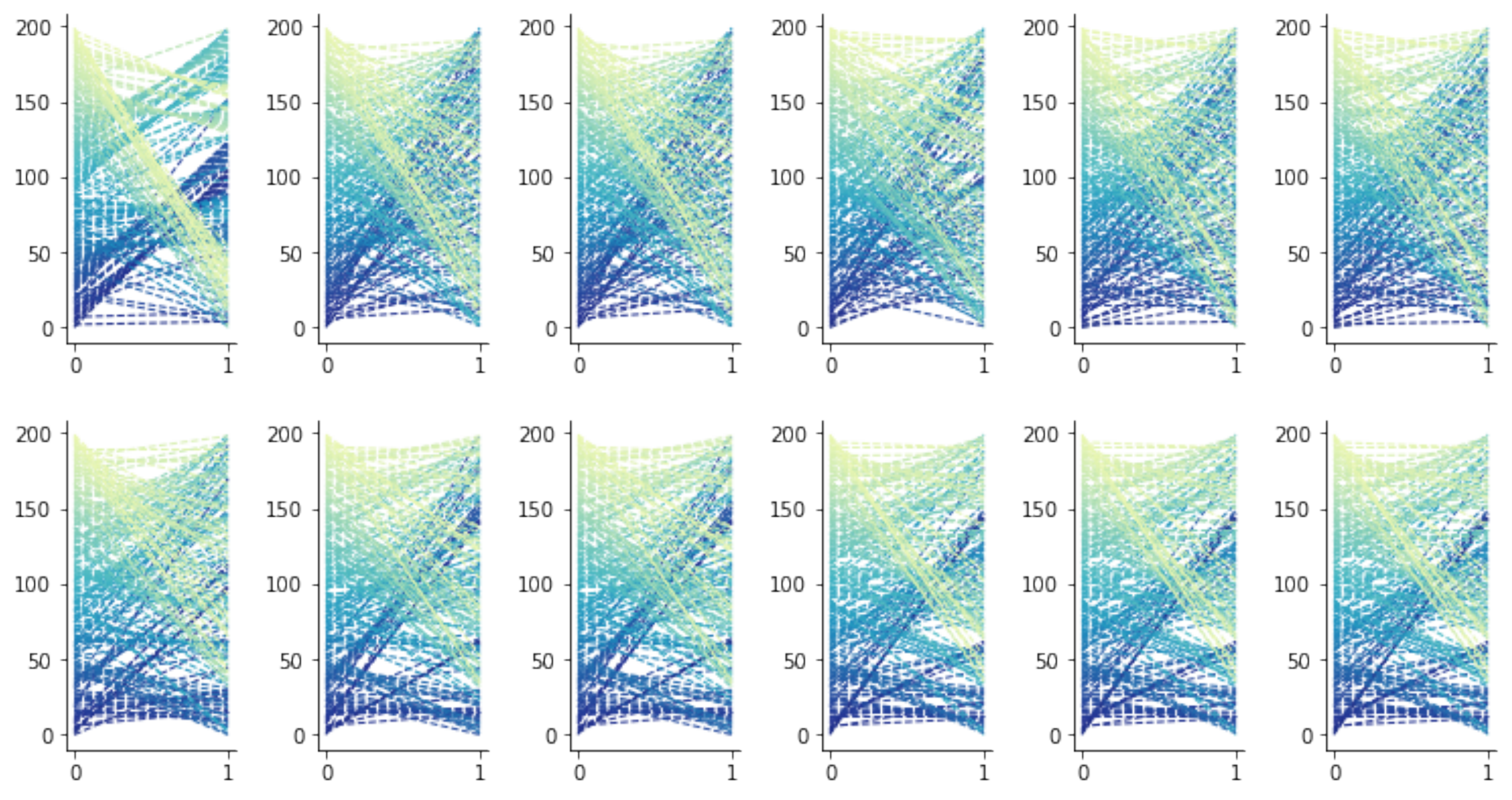}
    }
    \vspace{-0.5cm}
    \caption{
    \textbf{Ranking disorder examples.} We randomly select 12 runs from our experiments. For each sub-plot, 0 indicates the architecture ground-truth rank, and 1 indicates the ranking according to their super-net accuracy. We can clearly see that the ranking disorder happens uniformly across the search space and does not follow a particular pattern.
    }
    \label{fig:rank-disorder-example}
\end{figure}

\subsection{Limitation of Sparse Kendall-Tau}
\label{apdx:limitation}
We nonetheless acknowledge that our \skdt{} has some limitations. For example, a failure case of using \skdt{} for super-net evaluation may occur when the top 10\% architectures are perfectly ordered, while the bottom 90\% architectures are purely randomly distributed. In this case, the Kendall Tau will be close to 0. However, the search algorithm will always return the best model, as desired. 

Nevertheless, while this corner case would indeed be problematic for the standard Kendall Tau, it can be circumvented by tuning the threshold of our sKT. A large threshold value will lead to a small number of groups, whose ranking might be more meaningful. For instance in some randomly-picked NASBench-101 search processes, setting the threshold to 0.1\% merges the top 3000 models into 9 ranks, but still yields an sKT of only 0.2. Increasing the threshold to 10\% clusters the 423K models into 3 ranks, but still yields an sKT of only 0.3. This indicates the stability of our metric. 
In Figure~\ref{fig:rank-disorder-example}, we randomly picked 12 settings and show the corresponding bipartite graphs relating the super-net and ground-truth rankings to investigate where disorder occurs. In practice, the corner case discussed above virtually never occurs; the ranking disorder is typically spread uniformly across the architectures.

\section{Additional Results}

\subsection{Weight-sharing Protocol $P_{ws}$ -- Other Factors}
\label{apdx:wsp}

\mypara{Learning rate.}

\begin{figure}[!h]
    \centering
    \begin{minipage}{0.48\linewidth}
    \resizebox{\linewidth}{!}{
    \includegraphics{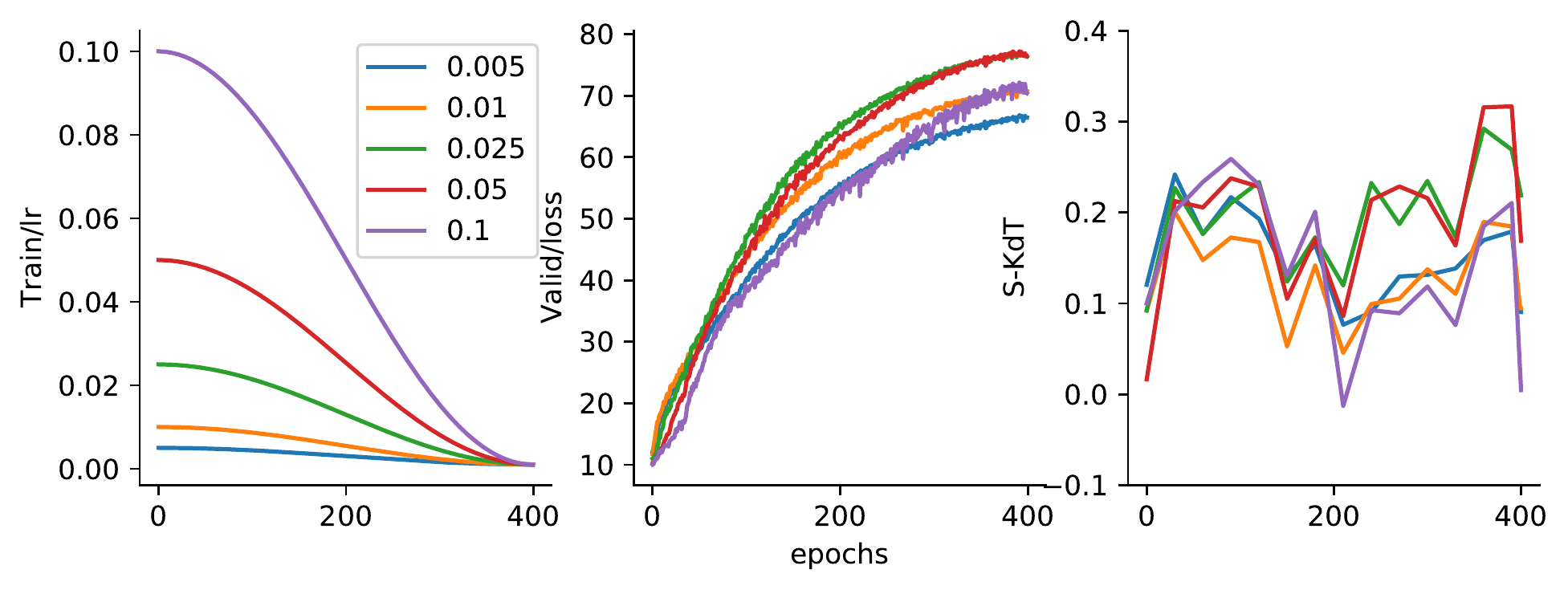}
    }
    \vspace{-0.7cm}
    \caption{\textbf{Learning rate} on NASBench-101.}
    \label{fig:supp:lr-nasbench101}
    \end{minipage}
    \hfill
    \begin{minipage}{0.48\linewidth}
    \resizebox{\linewidth}{!}{
    \includegraphics{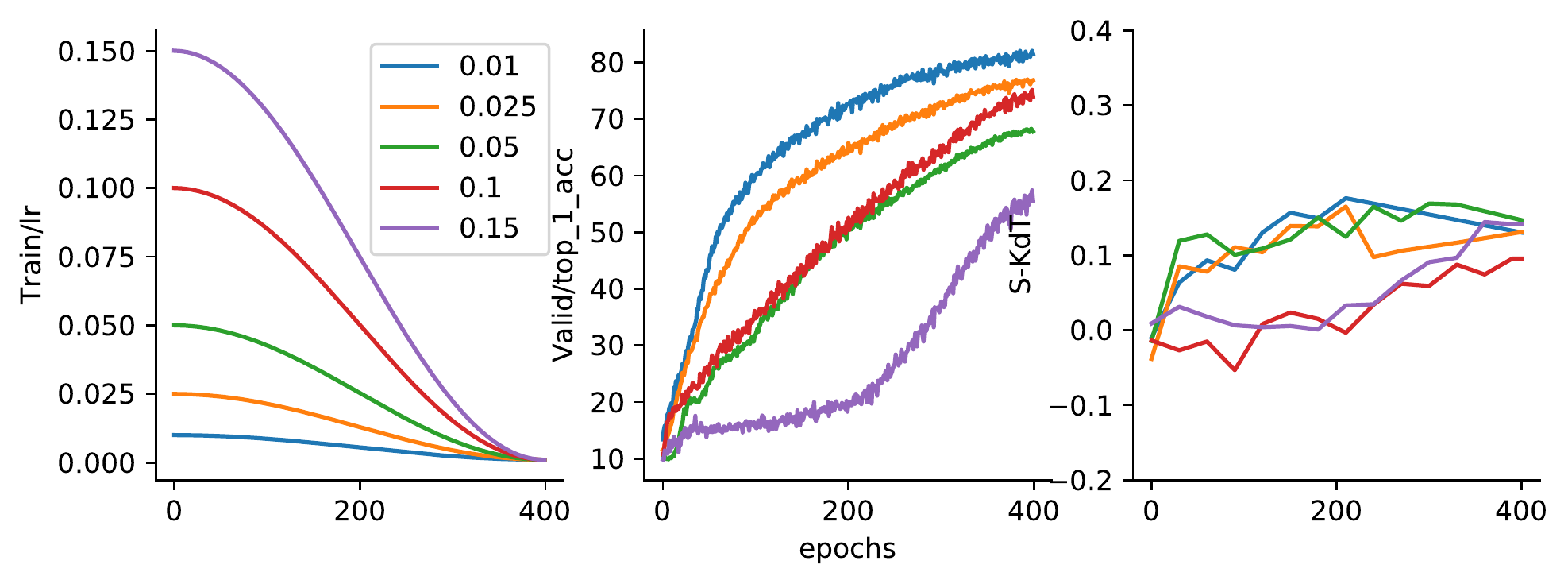}
    }
    \vspace{-0.7cm}
    \caption{\textbf{Learning rate} on DARTS-NDS.}
    \label{fig:supp:lr-darts-nds}
    \end{minipage}
\end{figure}
We report the learning rate validation results for NASBench-101 in Figure~\ref{fig:supp:lr-nasbench101} and for DARTS-NDS in Figure~\ref{fig:supp:lr-darts-nds}. For NASBench-101, we can see that learning rates of 0.025 and 0.05 clearly outperform 
other learning rates in terms of \skdt{} and validation accuracy. For DARTS-NDS, 
although the best validation accuracy is obtained with a learning rate of 0.01, the \skdt{} suggests that there is no significant difference once the learning rate is below 0.025, which is the stand-alone training learning rate. 
We pick 0.025 to be consistent with the other search spaces.

\mypara{Number of epochs.}

\begin{wrapfigure}{r}{0.5\linewidth}
\vspace{-0.5cm}
    \resizebox{\linewidth}{!}{
    \includegraphics{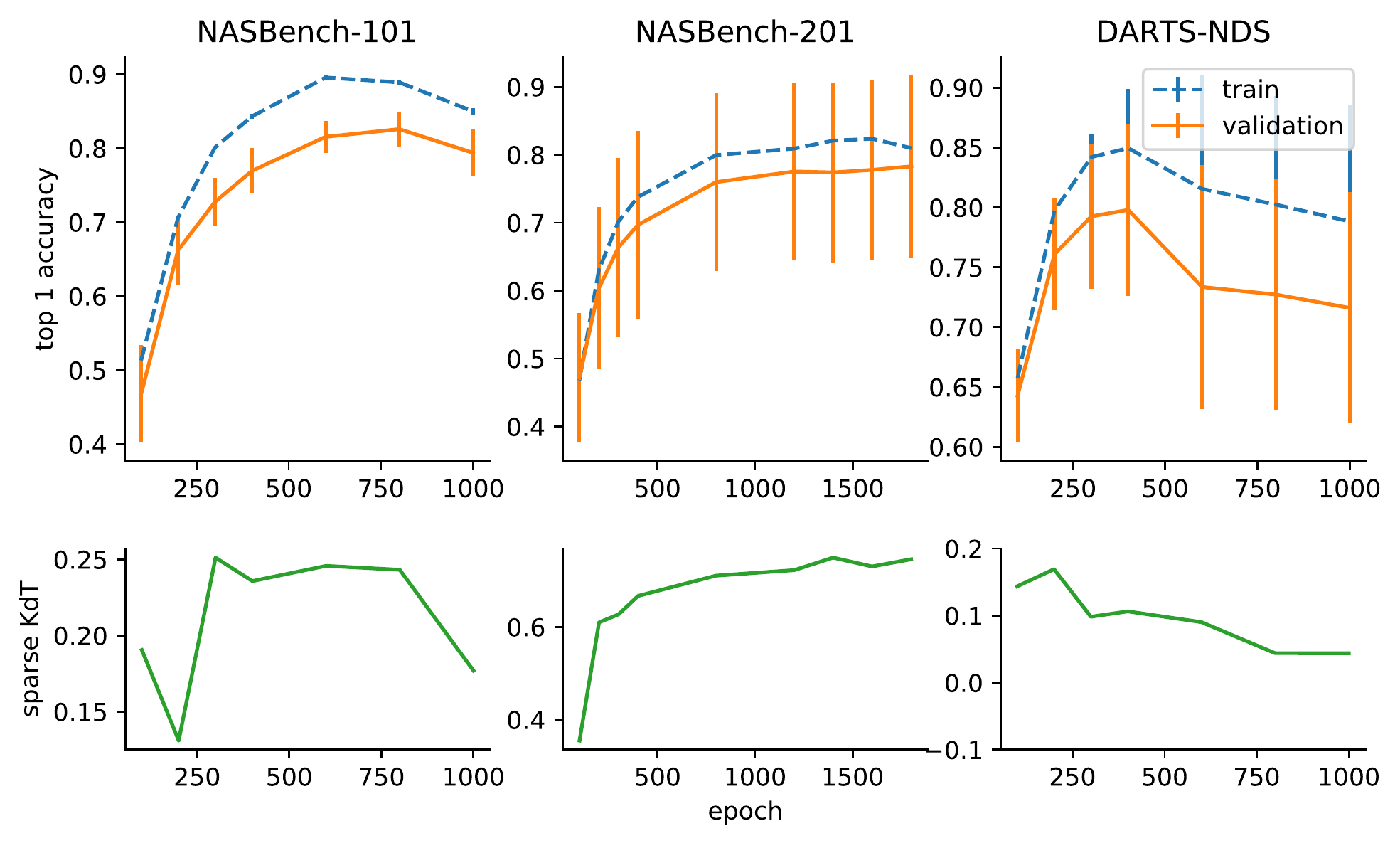}
    }
    \vspace{-0.7cm}
    \caption{\textbf{Validating the number of epochs.} 
    Each data point summarizes 3 individual runs. 
    }
    \label{fig:epochs}
    \vspace{-0.3cm}
\end{wrapfigure}
Since the cosine learning rate schedule decays the learning rate to zero towards the end of  training, we evaluate the impact of the number of training epochs. In stand-alone training, the number of epochs was set to 108 for NASBench-101, 200 for NASBench-201, and 100 for DARTS-NDS. Figure~\ref{fig:epochs} 
shows that increasing the number of epochs significantly improves the accuracy in the beginning, but eventually decreases the accuracy for NASBench-101 and DARTS-NDS. Interestingly, the number of epochs impacts neither the correlation of the ranking nor the final selected model performance after 400 epochs. 
We thus use 400 epochs for the remaining experiments.

\mypara{Weight decay.}

\begin{wrapfigure}{r}{0.5\linewidth}
    \centering
    \vspace{-0.5cm}
    \resizebox{\linewidth}{!}{
    \includegraphics{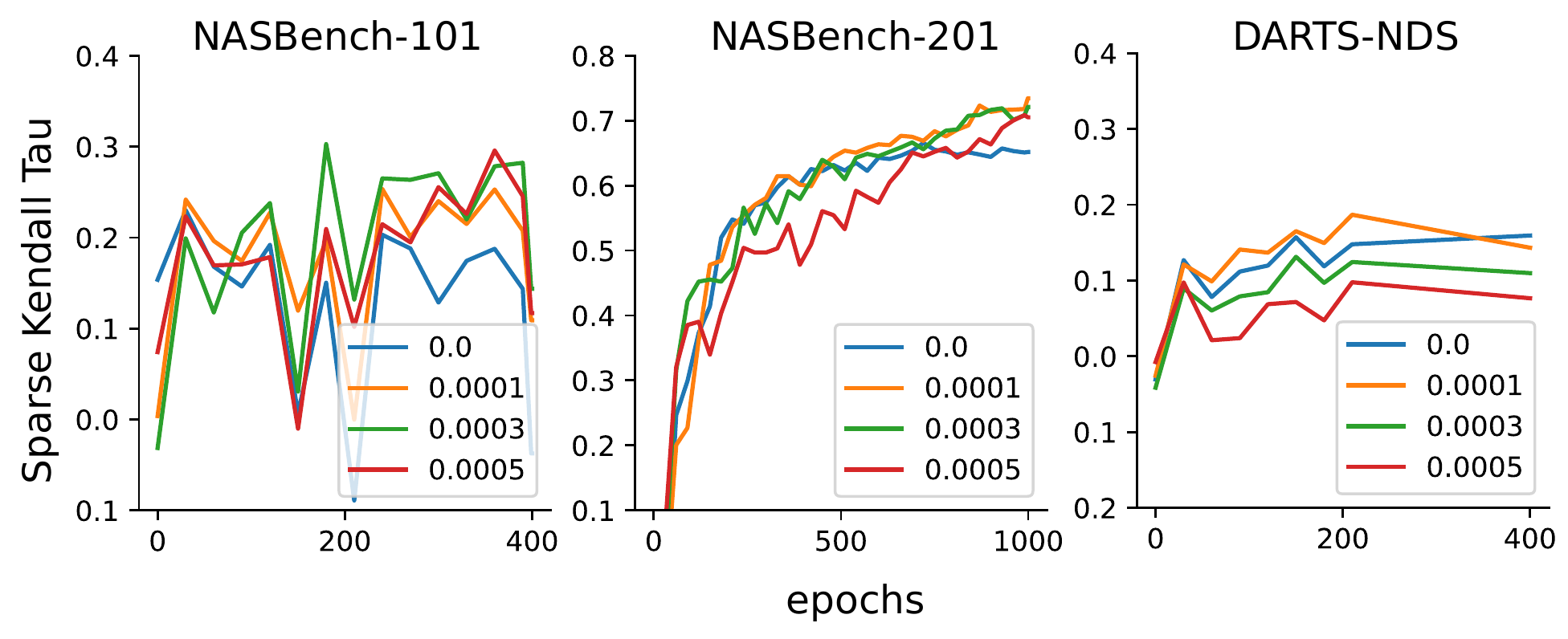}
    }
    \vspace{-0.6cm}
    \caption{\textbf{Weight decay validation.}
    }
    \vspace{-0.3cm}
    \label{fig:weight-decay}
\end{wrapfigure} 
Weight decay is used to reduce overfitting.
For WS-NAS, however, overfitting
does not occur because there are billions of architectures sharing the same set of parameters, which in fact rather causes underfitting.
Based on this observation,~\citet{nayman2019xnas} propose to disable weight decay during super-net training. Figure~\ref{fig:weight-decay}, however, shows that the behavior of weight decay varies across datasets. While on DARTS-NDS weight decay is indeed harmful, it improves the results on NASBench 101 and 201. 
We conjecture that this is due to the much larger number of architectures in DARTS-NDS (243 billion) than in the NASBench series (less than 500,000).

\begin{figure}
    \centering
    \begin{minipage}[t]{0.75\linewidth}
    \resizebox{\linewidth}{!}{
    \includegraphics{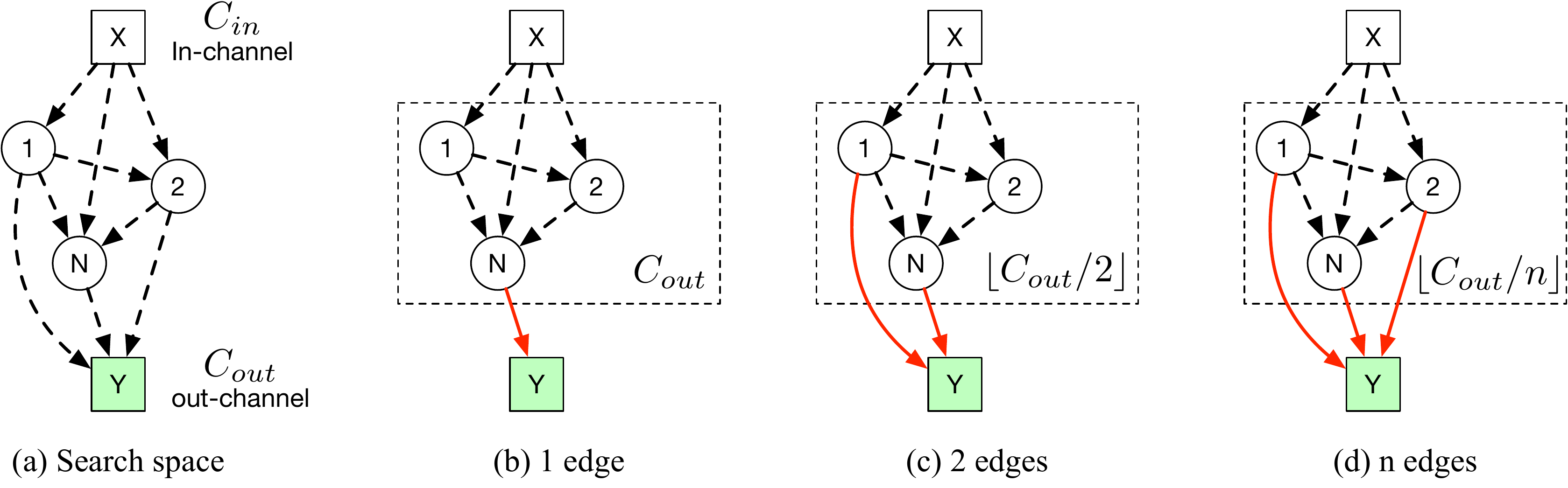}
    }
    \caption{\textbf{NASBench-101 dynamic channel}. \textbf{(a)} A search cell with $n$ intermediate nodes. $X$ and $Y$ are the input and output node with $C_{in}$ and $C_{out}$ channels, respectively.  \textbf{(b)} When there is only one incoming edge to the output node $Y$, all intermediate nodes will have $C_{out}$ channels. \textbf{(c)} When there are two edges, the associated channel numbers decrease
    so that the sum of the intermediate channel numbers equals $C_{out}$, i.e., the intermediate nodes have $\lfloor C_{out} / 2 \rfloor$ channels. 
     \textbf{(d)} In the general case, the intermediate channel number is $\lfloor C_{out} / n \rfloor $ for $n$ incoming edges.}
    \label{fig:disable-channel}
    \end{minipage}
    \hfill
    \begin{minipage}[t]{0.2\linewidth}
    \vspace{-2cm}
    \captionof{table}{NASBench-101 sub-spaces obtained by fixing the number of channels.}
    \resizebox{\linewidth}{!}{
    \begin{tabular}{c|c}
    \toprule
    \# Incoming     & \# Arch.  \\
     Edge &\\
    \midrule
    1 &  120933 \\
    2 & 201441 \\
    3 & 90782 \\
    4 & 10467 \\
    \bottomrule
    \end{tabular}
    }
    \label{tab:disable-channel}
    \end{minipage}
\end{figure}

\begin{wraptable}{r}{0.5\linewidth}
    \centering
    \vspace{-0.5cm}
    \caption{\textbf{Disabling dynamic channels}} 
    \resizebox{0.95\linewidth}{!}{
        \begin{tabular}{ l|cccc}
        \toprule
        Edges & Accuracy & S-KdT & P $>$ R & Final searched model \\
        \midrule
        \multicolumn{5}{l}{\textit{Baseline: random sampling sub-spaces with dynamic channeling.}} \\
1 &70.04$\pm$\phantom{0}8.15&0.173&0.797&91.19$\pm$2.01\\
2 &78.29$\pm$10.51&0.206&0.734&82.03$\pm$1.50\\
3 &79.92$\pm$\phantom{0}9.42&0.242&0.576&92.20$\pm$1.19\\
4 &79.37$\pm$\phantom{0}17.34&0.270&0.793&92.32$\pm$1.10\\
\midrule
Average       & 76.905 $\pm$ 10.05 & 0.223 & 0.865 & 89.435 $\pm$ 4.30 \\
        \midrule
        \midrule
        \multicolumn{5}{l}{\textit{Disable dynamic channels by fixing the edges to the output node.}} \\
1 &76.92$\pm$\phantom{0}7.87&0.435&0.991&93.94$\pm$0.22\\
2 &74.32$\pm$\phantom{0}8.21&0.426&0.925&93.34$\pm$0.01\\
3 &77.24$\pm$\phantom{0}9.18&0.487&0.901&93.66$\pm$0.07\\
4 &79.31$\pm$\phantom{0}7.04&0.493&0.978&93.65$\pm$0.07\\
\midrule
        Average       & 76.95 $\pm$ \phantom{0}8.29 & 0.460 & 0.949 & 93.65 $\pm$ 0.73 \\
        \bottomrule
        \end{tabular}
    }
    \vspace{-0.2cm}
    \label{tab:disable-dynamic-channel-results}
\end{wraptable}
\subsection{NASBench-101 -- Disabling Dynamic Channels}
\label{apdx:nasbench-101}
Here, we investigate how the dynamic channeling used in the NASBench 101 search space impacts the super-net quality and how to disable it. 
As shown in Figure~\ref{fig:disable-channel}, the number of channels of the intermediate layers depends on the number of incoming edges to the output node. A simple way to disable dynamic channelling then consists of separating the search space into multiple sub-spaces according to this criterion. Table~\ref{tab:disable-channel} indicates the number of architectures in each such sub-spaces. 

Since each sub-space now encompasses fewer architectures, it is not fair to perform a comparison with the full NASBench 101 search space. To this end, for each sub-space, we construct a baseline space where we drop architectures uniformly at random until the number of remaining architectures matches the size of the sub-space (cf. Table~\ref{tab:disable-channel}). We repeat this process with 3 different initializations, while keeping all other factors unchanged when training the super-net.

As shown in Table~\ref{tab:disable-dynamic-channel-results}, disabling dynamic channeling improves the \skdt{} and the final search performance by a large margin and yields a new state-of-the-art search performance on NASBench101. 

\subsection{Mapping \fws{} -- Other Factors}
\label{apdx:fws-other}
\begin{wraptable}{r}{0.5\linewidth}
\vspace{-0.5cm}
\caption{\textbf{Comparison of different mappings $f_{ws}$}. 
We report s-KdT $/$ final search performance.}
    \centering
    \resizebox{0.9\linewidth}{!}{
        \begin{tabular}{ l|ccc}
\toprule
& NASBench-101 & NASBench-201 & DARTS-NDS \\
\midrule
\midrule
Baseline          & 0.236 $/$ 92.32    & 0.740 $/$ 92.92    & 0.159 $/$ 93.59 \\
\midrule
WSBN              & 0.056 $/$ 91.33    & 0.675 $/$ 92.04  & 0.331 $/$ 92.95 \\
Global-Dropout    & 0.179 $/$ 90.95     & 0.676 $/$ 91.76   & 0.102 $/$ 92.30 \\
Path-Dropout      & 0.128 $/$ 91.19     & 0.431 $/$ 91.42   & 0.090 $/$ 91.90 \\
\midrule
OFA Kernel        & 0.132 $/$ 92.01    & 0.574 $/$ 91.83  & 0.112 $/$ 92.83 \\
\bottomrule
\end{tabular}
}
\label{tab:hp-fws}
\end{wraptable}
We evaluate the weight-sharing batch normalization~(WSBN) of \citet{Luo2018WSBNcode}
, which keeps an independent set of parameters for each incoming edge.
Furthermore, we test the two commonly-used dropout strategies: right before global pooling~(global dropout); and at all edge connections between the nodes~(path dropout). Note that path dropout has been widely used in WS-NAS~\cite{Luo2018,Liu2018darts,Pham2018}. For both dropout strategies, we set the dropout rate to 0.2. 
Finally, we evaluate the super convolution layer of~\cite{Cai2020Once}, referred to as OFA kernel, which accounts for the fact that, in CNN search spaces, convolution operations appear as groups, and thus merges the convolutions within the same group, keeping only the largest kernel parameters and performing a parametric projection to obtain the other kernels.
The results in Table~\ref{tab:hp-fws} 
show that all these factors negatively impact the search performances and the super-net quality.

\subsection{WS on Edges or Nodes?}
\label{apdx:node-edge}
\begin{wraptable}{r}{0.5\linewidth}
\vspace{-0.5cm}
\caption{\textbf{Comparison of operations on the nodes or on the edges.} 
We report sKT $/$ final search performance.}
    \centering
    \resizebox{0.9\linewidth}{!}{
        \begin{tabular}{ l|ccc}
\toprule
& NASBench-101 & NASBench-201 & DARTS-NDS \\
\midrule
\midrule
Baseline          & 0.236 $/$ 92.32    & 0.740 $/$ 92.92    & 0.159 $/$ 93.59 \\
\midrule
Op-Edge           & N/A & as Baseline   & 0.189 $/$ 93.97 \\
Op-Node           & as Baseline         & 0.738 $/$ 92.36   & as Baseline \\
\bottomrule
\end{tabular}
}
\vspace{0.2cm} 
\label{tab:edge-node}
\resizebox{0.9\linewidth}{!}{
\includegraphics[]{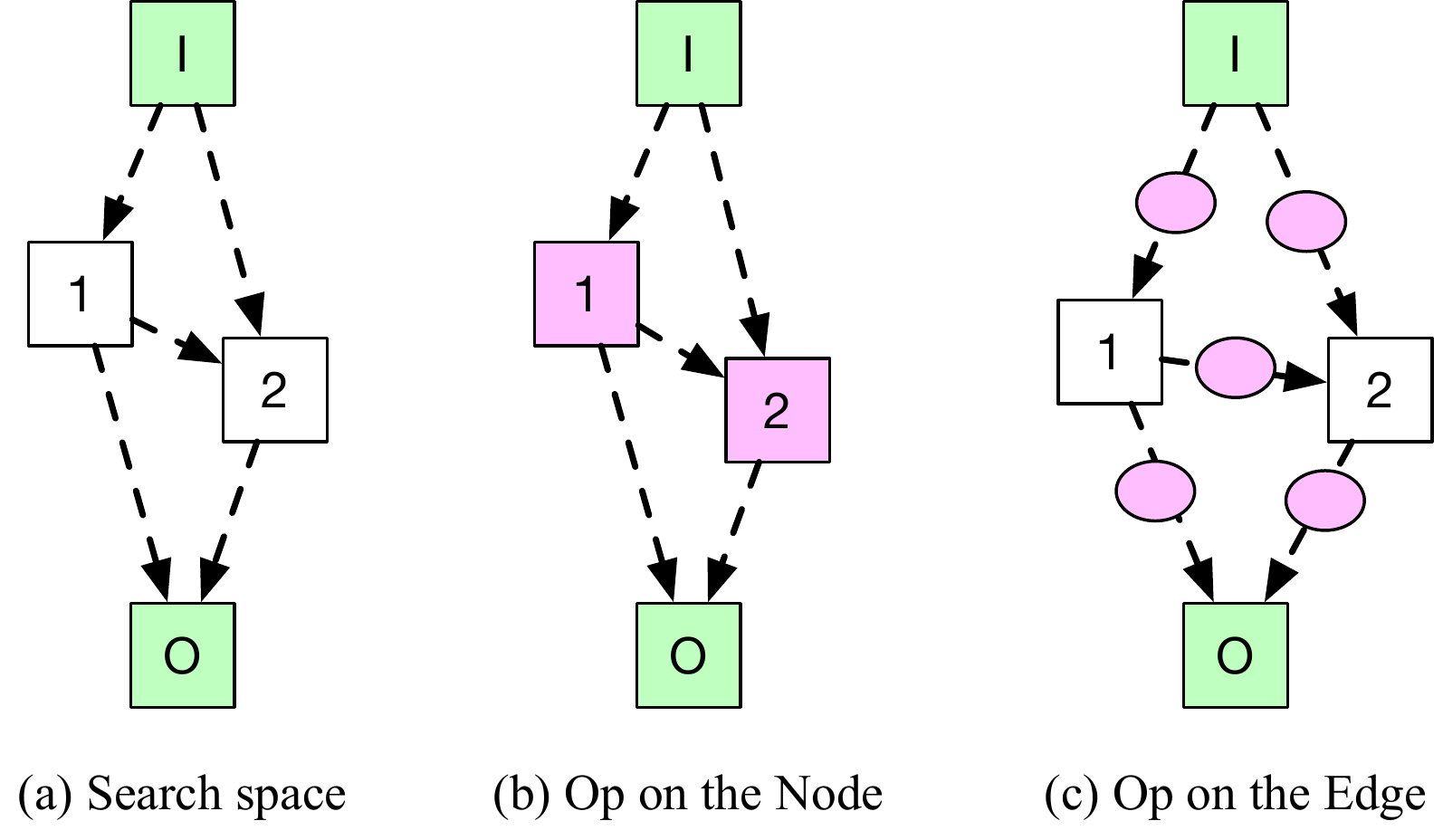}
}
\captionof{figure}{\textbf{(a)} Consider a search space with 2 intermediate nodes, 1, 2, with one input (I) and output (O) node. This yields 5 edges. Let us assume that we have 4 possible operations to choose from, as indicated as the purple color code. \textbf{(b)} When the operations are on the nodes, there are 2 $\times$ 4 ops to share, i.e., I$\rightarrow$2 and 1$\rightarrow$2 share weights on node 2. \textbf{(c)} If the operations are on the edges, then we have  5 $\times$ 4 ops to share.}
\label{fig:edge-node}
\end{wraptable}
Most existing works build $f_{ws}$ to define the shared operations on the graph nodes rather than on the edges. This is because, if $f_{ws}$ maps to the edges, the parameter size increases from $O(n)$ to $O(n^2)$, where $n$ is the number of intermediate nodes. %
We provide a concrete example in Figure~\ref{fig:edge-node}.
However, the high \skdt{} on NASBench-201 in the top part of Table~\ref{tab:edge-node}, which is obtained by mapping to the edges, may suggest that sharing on the edges is beneficial. Here we investigate if this is truly the case.

On NASBench-101, by design, each node merges the previous nodes' outputs and then applies parametric operations. This makes it impossible to build an equivalent sharing on the edges. We therefore construct sharing on the edges for DARTS-NDS and sharing on the nodes for NASBench-201. As shown in Table~\ref{tab:hp-fws}, for both spaces, sharing on the edges yields a marginally better super-net than sharing on the nodes. Such small differences might be due to the fact that, in both spaces, the number of nodes is 4, while the number of edges is 6, thus mapping to edges will not drastically affect the number of parameters. 
Nevertheless, this indicates that one should consider having a larger number of shared weights when the resources are not a bottleneck.

\subsection{Results for All Factors}
\label{apdx:allfactor}
We report the numerical results for all hyper-parameter factors in Table~\ref{tab:supp-huge-table-pws}, low-fidelity factors in Table~\ref{tab:supp-huge-table-low-fidelity} and implementation factors in Table~\ref{tab:supp-huge-table-implementation}. These results were computed from the last epochs of 3 different runs, as those reported in the main text. 
\begin{table}[!
h]
    \centering
    \caption{\textbf{Results for all WS Protocol $P_{ws}$ factors on the three search spaces.}} 
    \resizebox{\linewidth}{!}{
        \begin{tabular}{ l|cccc|cccc|cccc}
        \midrule[2pt]
        Factor & \multicolumn{4}{c}{NASBench-101} & \multicolumn{4}{c}{NASBench-201} & \multicolumn{4}{c}{DARTS-NDS} \\
        and & Super-net & \phantom{0} & \phantom{0} & Final & Super-net & \phantom{0} & \phantom{0} & Final & Super-net & \phantom{0} &  \phantom{0} & Final \\
        settings & Accuracy & S-KdT & P $>$ R &  Performance & Accuracy & S-KdT & P $>$ R & Performance & Accuracy & S-KdT & P $>$ R & Performance \\
        \midrule[1pt]
        \multicolumn{3}{l}{\textbf{Batch-norm. }} \\
        \midrule
affine F track F &0.651$\pm$0.05&0.161&0.996&0.916$\pm$0.13&0.660$\pm$0.13&0.783&0.997&92.67$\pm$1.21&0.735$\pm$0.18&0.056&0.224&93.14$\pm$0.28\\
affine T track F &0.710$\pm$0.04&0.240&0.996&0.924$\pm$0.01&0.713$\pm$0.14&0.718&0.707&91.71$\pm$1.05&0.265$\pm$0.21&-0.071&0.213&91.89$\pm$2.01\\
affine F track T &0.144$\pm$0.09&0.084&0.112&0.882$\pm$0.02&0.182$\pm$0.15&-0.171&0.583&86.41$\pm$4.84&0.359$\pm$0.25&-0.078&0.023&90.33$\pm$0.76\\
affine T track T &0.153$\pm$0.10&-0.008&0.229&0.905$\pm$0.01&0.134$\pm$0.09&-0.417&0.274&90.77$\pm$0.40&0.216$\pm$0.18&-0.050&0.109&90.49$\pm$0.32\\
        \midrule[1pt]
        \multicolumn{3}{l}{\textbf{Learning rate.}} \\
        \midrule
0.005 &0.627$\pm$0.07&0.091&0.326&0.908$\pm$0.01&0.658$\pm$0.11&0.668&0.141&90.14$\pm$0.55 & 0.792$\pm$0.08&0.130&0.033&91.81$\pm$0.68\\
0.01 &0.668$\pm$0.06&0.095&0.546&0.919$\pm$0.00&0.713$\pm$0.12&0.670&0.711&91.21$\pm$1.18 & 0.727$\pm$0.05&0.131&0.258&92.86$\pm$0.64\\
0.025 &0.715$\pm$0.05&0.220&0.910&0.917$\pm$0.01&0.659$\pm$0.13&0.665&0.844&92.42$\pm$0.58 & 0.656$\pm$0.14&0.218&0.299&93.42$\pm$0.20\\
0.05 &0.727$\pm$0.05&0.143&0.905&0.911$\pm$0.02&0.631$\pm$0.14&0.594&0.730&92.02$\pm$0.70&0.623$\pm$0.04&0.147&0.489&91.70$\pm$0.33\\
0.1 &0.690$\pm$0.07&0.005&0.905&0.909$\pm$0.02&0.609$\pm$0.28&0.571&0.618&91.82$\pm$0.81&0.735$\pm$0.06&0.096&0.099&92.73$\pm$0.24\\
0.15 &0.000$\pm$0.00&-0.274 & N/A & N/A &0.551$\pm$0.14&0.506&0.553&91.22$\pm$1.20&0.371$\pm$0.27&0.027&0.218&91.20$\pm$0.72\\
0.2 &-&-&-&-&0.519$\pm$0.12&0.557&0.035&88.74$\pm$0.11&0.102$\pm$0.48&-0.366&N/A&N/A\\
        \midrule[1pt]
        \multicolumn{10}{l}{\textbf{Epochs.}} \\
        \midrule
100 &0.468$\pm$0.07&0.190&0.759&0.920$\pm$0.01&0.472$\pm$0.09&0.355&0.997&92.11$\pm$1.67&0.643$\pm$0.04&0.144&0.901&93.90$\pm$0.49\\
200 &0.662$\pm$0.05&0.131&0.685&0.914$\pm$0.01&0.604$\pm$0.12&0.610&0.881&91.88$\pm$2.01&0.761$\pm$0.05&0.169&0.778&94.08$\pm$0.21\\
300 &0.727$\pm$0.03&0.251&0.739&0.920$\pm$0.01&0.664$\pm$0.13&0.627&0.840&91.42$\pm$1.91&0.793$\pm$0.06&0.098&0.870&93.22$\pm$0.95\\
400 &0.769$\pm$0.03&0.236&0.932&0.921$\pm$0.01&0.697$\pm$0.14&0.667&0.158&89.83$\pm$0.97&0.798$\pm$0.07&0.106&0.036&92.34$\pm$0.22\\
600 &0.815$\pm$0.02&0.246&0.556&0.911$\pm$0.01&0.720$\pm$0.13&0.682&0.285&90.28$\pm$0.82&0.734$\pm$0.10&0.090&0.209&93.23$\pm$0.19\\
800 &0.826$\pm$0.02&0.243&0.177&0.907$\pm$0.00&0.760$\pm$0.13&0.711&0.378&91.53$\pm$0.53&0.728$\pm$0.10&0.044&0.853&93.29$\pm$0.81\\
1000 &0.794$\pm$0.03&0.177&0.831&0.920$\pm$0.01&0.782$\pm$0.13&0.740&0.589&92.92$\pm$0.48&0.717$\pm$0.09&0.044&0.997&93.92$\pm$0.90\\
1200 &-&-&-&-&0.775$\pm$0.13&0.723&0.198&90.81$\pm$0.56&-&-&-&-\\
1400 &-&-&-&-&0.774$\pm$0.13&0.750&0.604&92.26$\pm$0.33&-&-&-&-\\
1600 &-&-&-&-&0.778$\pm$0.13&0.731&0.882&91.85$\pm$1.20&-&-&-&-\\
1800 &-&-&-&-&0.783$\pm$0.13&0.746&0.266&90.64$\pm$0.82&-&-&-&-\\
        \midrule[1pt]
        \multicolumn{10}{l}{\textbf{Weight decay.} } \\
        \midrule
0.0 &0.645$\pm$0.05&-0.037&0.179&0.899$\pm$0.01&0.713$\pm$0.13&0.652&0.266&90.58$\pm$0.99&0.670$\pm$0.03&0.159&0.629&93.09$\pm$0.73\\
0.0001 &0.719$\pm$0.03&0.109&0.659&0.912$\pm$0.01&0.756$\pm$0.13&0.734&0.612&91.88$\pm$0.59&0.751$\pm$0.05&0.143&0.396&93.37$\pm$0.44\\
0.0003 &0.771$\pm$0.03&0.144&0.648&0.915$\pm$0.01&0.772$\pm$0.13&0.721&0.726&92.34$\pm$0.57&0.759$\pm$0.06&0.110&0.890&93.82$\pm$0.51\\
0.0005 &0.782$\pm$0.03&0.117&0.910&0.911$\pm$0.02&0.764$\pm$0.13&0.705&0.882&92.61$\pm$0.59&0.739$\pm$0.07&0.077&0.051&91.61$\pm$1.01\\
        \midrule[1pt]
\multicolumn{10}{l}{\textbf{Sampling.} } \\
\midrule
Random-A &0.717$\pm$0.04&0.133&0.862&0.919$\pm$0.02&0.764$\pm$0.13&0.705&0.882&92.61$\pm$0.59&-&-&-&-\\
Random-NAS &0.638$\pm$0.20&0.167&0.949&0.913$\pm$0.02&0.765$\pm$0.14&0.750&0.897&92.17$\pm$1.01&-&-&-&-\\
FairNAS &0.789$\pm$0.03&0.288&0.382&0.908$\pm$0.01&0.774$\pm$0.14&0.713&0.917&93.06$\pm$0.31&-&-&-&-\\
        \midrule[2pt]
        \end{tabular}
    }
    \label{tab:supp-huge-table-pws}
	\caption{\textbf{Results for all low-fidelity factors on the three search spaces.}} 
\resizebox{\linewidth}{!}{
	\begin{tabular}{ l|cccc|cccc|cccc}
        \midrule[2pt]
		Factor & \multicolumn{4}{c}{NASBench-101} & \multicolumn{4}{c}{NASBench-201} & \multicolumn{4}{c}{DARTS-NDS} \\
		and & Super-net & \phantom{0} & \phantom{0} & Final & Super-net & \phantom{0} & \phantom{0} & Final & Super-net & \phantom{0} &  \phantom{0} & Final \\
		settings & Accuracy & S-KdT & P $>$ R &  Performance & Accuracy & S-KdT & P $>$ R & Performance & Accuracy & S-KdT & P $>$ R & Performance \\
        \midrule[1pt]
		\multicolumn{10}{l}{\textbf{Number of Layer} (-X indicates the baseline minus X)} \\
		\midrule
		Baseline & 0.769$\pm$0.03&0.236&0.932&0.921$\pm$0.01 & 0.782$\pm$0.13&0.740&0.589&92.92$\pm$0.48 & 0.670$\pm$0.03&0.159&0.629&93.09$\pm$0.73 \\
		-1 &0.759$\pm$0.03&0.214&0.222&0.901$\pm$0.01 & 0.749$\pm$0.13&0.710&0.796&91.85$\pm$0.92 &0.843$\pm$0.04&0.178&0.299&92.35$\pm$1.25 \\
		-2 & 0.817$\pm$0.03&0.228&0.713&0.910$\pm$0.02 & 0.777$\pm$0.13&0.700&0.822&92.68$\pm$0.37 & 0.852$\pm$0.03&0.205&0.609&92.65$\pm$1.89\\
        \midrule[1pt]
		\multicolumn{3}{l}{\textbf{Train portion}} \\
		\midrule
		0.25 &0.433$\pm$0.07&0.216&0.281&0.901$\pm$0.01&0.660$\pm$0.11&0.668&0.979&92.30$\pm$1.14&0.597$\pm$0.14&0.132&0.359&92.27$\pm$1.84\\
		0.5 &0.612$\pm$0.06&0.251&0.424&0.896$\pm$0.02&0.740$\pm$0.12&0.669&0.979&93.17$\pm$0.47&0.666$\pm$0.17&0.083&0.551&92.22$\pm$1.36\\
		0.75 &0.688$\pm$0.05&0.222&0.857&0.920$\pm$0.01&0.758$\pm$0.13&0.725&0.618&92.46$\pm$0.19&0.715$\pm$0.18&0.096&0.081&92.29$\pm$0.47\\
		0.9 &0.722$\pm$0.05&0.186&0.996&0.931$\pm$0.01&0.772$\pm$0.13&0.721&0.726&92.34$\pm$0.57&0.703$\pm$0.18&0.042&0.065&92.78$\pm$0.10\\
        \midrule[1pt]
		\multicolumn{10}{l}{\textbf{Batch size} (/ X indicates the baseline divide by X)} \\
		\midrule
		Baseline & 0.769$\pm$0.03&0.236&0.932&0.921$\pm$0.01 & 0.782$\pm$0.13&0.740&0.589&92.92$\pm$0.48 & 0.670$\pm$0.03&0.159&0.629&93.09$\pm$0.73 \\
		/ 2 & 0.670$\pm$0.05&0.246&0.807&0.920$\pm$0.01 & 0.728$\pm$0.16&0.719&0.842&92.37$\pm$0.61 &0.698$\pm$0.20&0.037&0.209&93.24$\pm$0.13\\
		/ 4 &0.686$\pm$0.07&0.155&0.913&0.921$\pm$0.01&0.703$\pm$0.16&0.679&0.672&92.35$\pm$0.34 & 0.633$\pm$0.20&0.033&0.690&93.68$\pm$0.62\\
        \midrule[1pt]
		\multicolumn{10}{l}{\textbf{\# channel} (/ X indicates the baseline divide by X)} \\
		\midrule
		Baseline & 0.769$\pm$0.03&0.236&0.932&0.921$\pm$0.01 & 0.782$\pm$0.13&0.740&0.589&92.92$\pm$0.48 & 0.670$\pm$0.03&0.159&0.629&93.09$\pm$0.73 \\
		/ 2 &0.658$\pm$0.05&0.156&0.704&0.898$\pm$0.02 & 0.697$\pm$0.14&0.667&0.158&89.83$\pm$0.97 & 0.776$\pm$0.05&0.190&0.993&93.90$\pm$0.71\\
		/ 4 &0.604$\pm$0.06&0.093&0.907&0.922$\pm$0.01& 0.606$\pm$0.13&0.616&0.878&92.86$\pm$0.34 &0.707$\pm$0.05&0.202&0.359&92.93$\pm$0.58 \\
        \midrule[2pt]
	\end{tabular}
}
\label{tab:supp-huge-table-low-fidelity}

	\caption{\textbf{Results for all implementation factors on the three search spaces.}} 
\resizebox{\linewidth}{!}{
	\begin{tabular}{ l|cccc|cccc|cccc}
        \midrule[2pt]
		Factor & \multicolumn{4}{c}{NASBench-101} & \multicolumn{4}{c}{NASBench-201} & \multicolumn{4}{c}{DARTS-NDS} \\
		and & Super-net & \phantom{0} & \phantom{0} & Final & Super-net & \phantom{0} & \phantom{0} & Final & Super-net & \phantom{0} &  \phantom{0} & Final \\
		settings & Accuracy & S-KdT & P $>$ R &  Performance & Accuracy & S-KdT & P $>$ R & Performance & Accuracy & S-KdT & P $>$ R & Performance \\
        \midrule[1pt]
		\multicolumn{10}{l}{\textbf{Other factors}} \\
		\midrule
				Baseline & 0.769$\pm$0.03&0.236&0.932&0.921$\pm$0.01 & 0.782$\pm$0.13&0.740&0.589&92.92$\pm$0.48 & 0.670$\pm$0.03&0.159&0.629&93.09$\pm$0.73 \\
		OFA Kernel & 0.708$\pm$0.08 & 0.132 & 0.203 & 92.01$\pm$0.19 & 0.672$\pm$0.18 &0.574 & 0.605 & 91.83 $\pm$ 0.86& 0.782$\pm$0.05&0.112&0.399&93.22$\pm$0.43 \\
		WSBN  &0.155$\pm$0.07&0.085&0.504&0.809$\pm$0.13&0.703$\pm$0.14&0.676&0.585&92.06$\pm$0.48&0.744$\pm$0.16&0.033&0.682&92.88$\pm$1.22\\
        \midrule[1pt]
		\multicolumn{10}{l}{\textbf{Path dropout rate}} \\
		\midrule
			Baseline & 0.769$\pm$0.03&0.236&0.932&0.921$\pm$0.01 & 0.782$\pm$0.13&0.740&0.589&92.92$\pm$0.48 & 0.670$\pm$0.03&0.159&0.629&93.09$\pm$0.73 \\
		0.05 &0.750$\pm$0.02&0.206&0.819&0.915$\pm$0.07 &0.490$\pm$0.09&0.712&0.881&92.25$\pm$0.89&0.184$\pm$0.06&0.006&0.359&92.93$\pm$0.60\\
		0.15 &0.726$\pm$0.02&0.186&0.482&0.910$\pm$0.01&0.250$\pm$0.03&0.640&0.526&91.44$\pm$1.25&0.366$\pm$0.05&0.059&0.570&92.61$\pm$1.28\\
		0.2 &0.669$\pm$0.01&0.110&0.282&0.901$\pm$0.01&0.185$\pm$0.02&0.431&0.809&92.15$\pm$0.85&0.518$\pm$0.06&0.090&0.009&91.45$\pm$0.58\\
        \midrule[1pt]
		\multicolumn{10}{l}{\textbf{Global dropout}} \\
		\midrule
		Baseline & 0.769$\pm$0.03&0.236&0.932&0.921$\pm$0.01 & 0.782$\pm$0.13&0.740&0.589&92.92$\pm$0.48 & 0.670$\pm$0.03&0.159&0.629&93.09$\pm$0.73 \\
		0.2 &0.739$\pm$0.05&0.233&0.221&0.910$\pm$0.00&0.712$\pm$0.13&0.702&0.950&91.76$\pm$1.36&0.557$\pm$0.19&0.018&0.451&93.51$\pm$0.27\\
        \midrule[2pt]
       \multicolumn{10}{l}{Please refer to Appendix~\ref{apdx:node-edge} for mapping on the node or edge and Appendix~\ref{apdx:nasbench-101} for dynamic channel factor results.}
	\end{tabular}
}
\vspace{-0.4cm}
\label{tab:supp-huge-table-implementation}
\end{table}

\end{document}